\documentclass{article}

\PassOptionsToPackage{table}{xcolor}
\usepackage{iclr2027_conference,times}

\AtBeginDocument{\let\cite\citep}

\iclrfinalcopy

\usepackage[utf8]{inputenc}
\usepackage[T1]{fontenc}
\usepackage{hyperref}
\usepackage{url}
\usepackage{booktabs}
\usepackage{amsfonts}
\usepackage{nicefrac}
\usepackage{microtype}
\usepackage{xcolor}
\usepackage{algorithm}
\usepackage{algpseudocode}
\usepackage{xspace}
\usepackage{booktabs}
\usepackage{graphicx}
\usepackage{amsmath}
\usepackage{amssymb}
\usepackage{tikz} 
\usepackage{graphicx} 
\usepackage{caption} 
\usepackage{makecell}
\usepackage{multirow}
\usepackage{array}
\usepackage{cleveref}
\usepackage{subcaption}
\usepackage{sidecap}
\usepackage{enumitem}
\usepackage{float}
\usepackage[utf8]{inputenc}
\usepackage[T1]{fontenc}
\usepackage{booktabs}   
\usepackage{amsmath}
\usepackage{microtype}  
\usepackage{tabularx}   
\usepackage{makecell}
\usepackage{pifont}
\usepackage{xspace}
\usepackage[bottom,belowfloats]{footmisc}
\newcommand{\oursmethod}{\quad + \textbf{\textit{BTC3D}}}

\newcommand{\metric}[2]{#1 $#2$}

\newcommand{\minisection}[1]{\vspace{0.02in}\noindent{\bf #1}}

\newcommand{\ourmethod}{\textit{\textbf{BTC3D}}\xspace}

\newcommand{\btcembed}{\textit{\textbf{BTCemb}}\xspace}
\newcommand{\dycond}{\textit{\textbf{DyCond}}\xspace}
\newcommand{\cmark}{\ding{51}\xspace}%
\newcommand{\xmark}{\ding{55}\xspace}%
\usepackage{soul}

\title{BTC3D: Blended Tile Conditioning for \\ Detail-Enhancing Image-to-3D Generation}

\author{
\vspace{-3mm}
{\normalfont
\begin{tabular}{@{}l@{}}
\textbf{Junyu Li}$^{1}$,
\textbf{Qiuyu Chen}$^{1}$,
\textbf{Pengcheng Wang}$^{1}$,
\textbf{Shiqi Yang}$^{2}$,
\\
\textbf{Alexandra Gomez-Villa}$^{3}$,
\textbf{Joost van de Weijer}$^{3}$,
\textbf{Ruilin Li}$^{1,\dagger}$,
\textbf{Kai Wang}$^{1,\dagger}$
\\[0.4ex]
$^{1}$City University of Hong Kong (Dongguan),
$^{2}$SB Intuitions Corp.,
\\
$^{3}$Universitat Autònoma de Barcelona
\\[0.2ex]
\texttt{\{junyu.li, qiuyu.chen, pengcheng.wang\}@cityu-dg.edu.cn}
\\
\texttt{\{ruilin.li, kai.wang\}@cityu-dg.edu.cn}
\\
\texttt{shiqi.yang147.jp@gmail.com},
\texttt{\{agomezvi, joost\}@cvc.uab.cat}
\end{tabular}
}
}

\begin{document}

\maketitle

\begin{abstract}
\vspace{-3mm}
Recent diffusion-based pipelines have achieved promising progress in image-to-3D synthesis. 
However, generating high-fidelity details remains challenging, especially when the input image contains rich details. 
Existing approaches often rely on globally encoded conditioning features, which compress spatial information and limit the model to reproduce fine-grained details. 
This common design often leads to a phenomenon we term \textit{detail attenuation}. 
Moreover, improving image-to-3D synthesis quality typically requires retraining or fine-tuning large diffusion models, which can be computationally expensive and impractical for complex 3D pipelines.
In this work, we present \textbf{B}lended \textbf{T}ile \textbf{C}onditioning for image-to-\textbf{3D} generation (\ourmethod), a \textit{training-free} inference time framework that enhances fine-grained detail preservation in image-to-3D diffusion pipelines.
To alleviate \textit{detail attenuation}, we first examine the \textit{image feature additivity} in image-to-3D models.
Based on this property, we introduce a \textit{blended tile embedding} that extracts local conditioning signals from split image regional patches, allowing the diffusion model to better preserve fine-grained visual details. 
To integrate the global and local conditioning guidance stably, we propose a \textit{dynamic conditioning} schedule that gradually increases the influence of tile-level conditioning during later low-noise stages of diffusion.
Our proposed method \ourmethod operates entirely at inference time and can be seamlessly integrated into existing image-to-3D diffusion pipelines. 
Experimental results demonstrate that the proposed approach significantly improves texture quality and visual fidelity of the base model while maintaining global structural consistency in a training-free manner.

\begin{figure}[!b]
    \centering
    \vspace{-5mm}
    \includegraphics[width=0.95\linewidth]{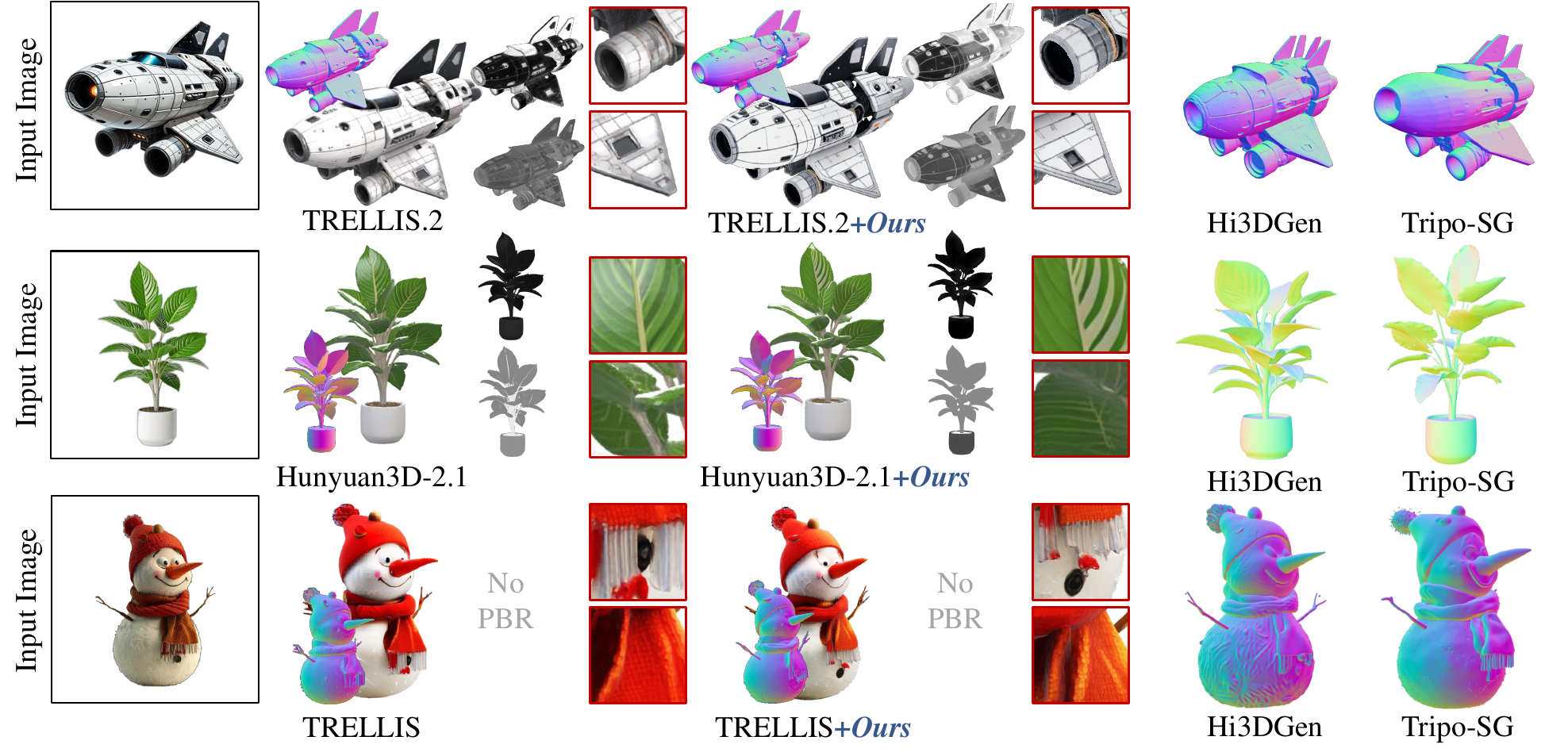}
    \vspace{-4mm}
    \caption{
    Baselines vs. \ourmethod. The image-to-3D models suffer from \textit{detail attenuation} in 3D synthesis.
    The details from the input conditional images are not recognized in the 3D assets.
    By comparison, applying our \ourmethod to TRELLIS/TRELLIS.2/Hunyuan3D-v2.1 augments the 3D assets generation by enriching such detail preservation.
    }
    \vspace{-1mm}
    \label{fig:teaser}
\end{figure}

\end{abstract}
\lhead{Preprint}
\renewcommand{\thefootnote}{\fnsymbol{footnote}}
\footnotetext[2]{Corresponding authors.}
\renewcommand{\thefootnote}{\arabic{footnote}}

\section{Introduction}
\label{sec:intro}

Generative models for 3D asset creation have witnessed significant advancements ~\cite{li2025step1x_3d,li2025triposg,wu2025direct3d,ye2025hi3dgen}, facilitating their widespread adoption across diverse domains, including entertainment~\cite{xu2024sketch2scene_3dgame}, robotics~\cite{ke20243d_robot} and healthcare~\cite{khader2023denoising_3dmedical}.
Recently, flow matching techniques~\cite{lipman2022flow_matching,liu2022_rectified_flow} have been applied to image-to-3D synthesis, exemplified by TRELLIS~\cite{xiang2025trellis2,xiang2025structured_trellis} and Hunyuan3D~\cite{yang2024hunyuan3d_v1,hunyuan3d22025tencent_v2} series. 
They demonstrate strong ability to generate high-fidelity 3D objects by mapping heterogeneous representations into a unified latent space, while explicitly disentangling geometric structures from appearance attributes.
Despite their success, generating high-quality textures remains a challenging problem for image-to-3D models, which we refer to as \textit{detail attenuation}. 
This limitation becomes particularly evident when preserving detailed textures, as shown in~\Cref{fig:teaser}.

The \textit{detail attenuation} results from a common design in existing pipelines~\cite{li2025triposg,hunyuan3d22025tencent_v2,hunyuan3d2025hunyuan3d_v21,xiang2025trellis2,xiang2025structured_trellis}, where these 3D generative models often encode the entire input image into a single global conditioning representation. 
While this representation captures coarse-grained semantic information, it inevitably compresses spatial details and reduces the model's ability to reproduce fine-grained textures. 
As a result, generated textures often appear over-smoothed or lack details.
Another limitation lies in the diffusion sampling process itself. 
During the denoising stage, early timesteps mainly determine global structure, whereas generated details are formed during the late stages of sampling. 
However, standard pipelines use the same conditioning representation throughout the sampling trajectory, despite the different roles of early and late generation stages.

To address \textit{detail attenuation}, we investigate whether generation quality can be improved at inference time without modifying model parameters. Our key insight is that global and local image conditioning serve different roles during the coarse-to-fine generation process: global conditioning is important for overall structure during the early generation steps, while local conditioning provides fine-grained cues that become more useful as generation progresses toward detail synthesis.  Combining global and local conditioning is enabled by an empirical property we identify in image-to-3D models: \textit{image feature additivity}. Specifically, we find that combining feature-level representations of local image patches produces semantically aligned 3D outputs, preserving spatially localized details that global conditioning alone tends to suppress. This property is consistent with observations in 2D image models~\cite{girdhar2023imagebind,qin2025free_sadis} and language models~\cite{hu2024token_merging_tome,mikolov2013efficient_word2vec}, but has not been explored in the context of 3D generation.

Motivated by this finding, we introduce \textbf{B}lended \textbf{T}ile \textbf{C}onditioning (\ourmethod), which extracts localized conditioning features from image patches and aggregates them via a foreground-based weighting scheme into a \textit{Blended Tile Conditioning Embedding} (\btcembed). This embedding complements the global conditioning signal, providing the model with fine-grained local cues that improve detail preservation. To integrate these two levels of conditioning stably, we further propose a \textit{dynamic conditioning} (\dycond) mechanism that progressively increases the influence of \btcembed during the later diffusion stage, where fine details are formed, while maintaining global structural consistency.

To verify the effectiveness of our method \ourmethod, we evaluate existing image-to-3D models against our training-free framework on the public 3D-Arena~\citep{3d-arena} and Toys4K~\citep{stojanov2021using_toys4k} benchmarks.
From both quantitative and qualitative experimental results, \ourmethod demonstrates improvements while applied to existing image-to-3D pipelines without retraining. To summarize, our contributions are as follows:

\begin{itemize}
    \item We identify the \textit{detail attenuation} problem in existing image-to-3D models. To alleviate the detail attenuation problem, we are the first to reveal the \textit{image feature additivity} property in image-to-3D models, which is the underlying mechanism of our training-free framework. 
    
    \item We introduce a \textit{training-free} mechanism named \textit{blended tile conditioning} (\ourmethod), which extracts localized conditioning feature \btcembed from regional image patches. 
    It is further refined by exploiting the low-noise regime for 3D texture enhancement.
    We thus design a \textit{dynamic conditioning} schedule that progressively integrates local guidance during diffusion, improving visual fidelity while maintaining structural consistency.
    
    \item Extensive experiments on multiple benchmarks demonstrate that \ourmethod achieves state-of-the-art performance in image-to-3D generation. Moreover, it can be seamlessly integrated into existing diffusion-based image-to-3D pipelines without retraining.

\end{itemize}

\section{Related Works}
Two dominant paradigms have been established for high-quality image-to-3D generation. The first one lifts 2D diffusion models to 3D by optimization-based distillation and the second one, named native 3D generation paradigm, enables precise geometry synthesis by explicit geometric representation modeling and learned 3D feature extraction.

\minisection{3D Generation via 2D priors.}
The first prevalent branch focuses on distilling useful priors from pretrained 2D generative models~\cite{podell2023sdxl,ramesh2022dalle2,Rombach_2022_CVPR_stablediffusion} into feed-forward 3D reconstruction algorithms, which can be divided into data-based and gradient-based distillation strategies.
Data distillation fine-tunes 2D models to synthesize multi-view images~\cite{shi2023mvdream,shriram2024realmdreamer,yu2024viewcrafter}, which are further reconstructed into 3D assets via techniques such as 3D Gaussian splatting~\cite{kerbl2023_3dgs}.
Gradient distillation, typified by Score Distillation Sampling introduced by the pivotal work DreamFusion~\cite{poole2023dreamfusion}, directly guides 3D optimization through gradient signals from 2D diffusion models. This technique was followed by numerous successors~\cite{tang2023dreamgaussian,wang2024luciddreaming,wang2023prolificdreamer}.
Despite their effectiveness, distillation-based approaches lack an explicit latent 3D space, which severely limits fine-grained structural control over the generated assets.
They especially struggle at keeping multi-view consistency and the precise alignment of texture with fine-grained geometric details.
These problems could hardly be avoided as they are deeply rooted in the inherent limitations of the distillation pipeline itself, unless one can generate textures directly and natively within 3D space.

\minisection{Native 3D Generation.}
To overcome the above limitations, the second paradigm pursues unified native 3D generative models trained from scratch.
Inspired by the remarkable success of diffusion models in 2D image and video synthesis~\cite{ho2020ddpm,khachatryan2023text2video}, recent methods~\cite{cheng2023sdfusion,ren2024xcube,vahdat2022lion,zeng2024paint3d} learn compact latent representations and perform generation within this compressed latent space, significantly advancing 3D generative modeling.
Benefiting from large-scale 3D datasets~\cite{deitke2023objaverse_plus,deitke2023objaverse}, modern 3D foundation models~\cite{chen20253dtopia,hong2023lrm,li2025triposg,hunyuan3d2025hunyuan3d_v21} have been developed to capture strong geometric priors, enabling not only high-quality generation across diverse object categories but also various downstream applications such as shape analysis~\cite{du2025hierarchical} and 3D editing~\cite{hu2024neural,li2025voxhammer}.
Notably, image-to-3D synthesis currently achieves substantially higher fidelity and controllability than text-to-3D synthesis.
However, most existing native 3D generators are restricted to either explicit representations (e.g., point clouds, voxels, meshes) or implicit formulations (e.g., neural fields, 3D Gaussians).
TRELLIS~\cite{xiang2025structured_trellis} addresses this constraint by introducing a Structured Latent Representation (SLAT) that supports flexible multi-format 3D generation, whose versatility we inherit to accommodate diverse 3D modalities in our framework.
Although these 3D foundation models exhibit strong generalization abilities, they mainly generate samples from a learned data distribution and lack dedicated mechanisms for user-specified personalization or example-driven control.
In this paper, we leverage the powerful priors encoded in the 3D foundation models and introduce a training-free refinement framework.

\section{Methodology}
\label{sec:method}

To address the \textit{detail attenuation} problem in image-to-3D generations, we propose a training-free, inference-time framework, named \textit{Blended Tile Conditioning} for image-to-3D generation (\ourmethod). 
\ourmethod builds on the key insight that global conditioning is more important early in generation, while local conditioning becomes more useful for fine details at later stages.
In~\Cref{subsec:observation}, we demonstrate the empirical basis of our method, the compositionality and compatibility of image features in generation. 
Based on these observed properties, we then build our approach in~\Cref{subsec:overview} and illustrate it in~\Cref{fig:method}.

\subsection{Empirical Observations on DINO-encoded Features}
\label{subsec:observation}

\textbf{Task Definition.}
Image-to-3D generation aims to recover a complete 3D representation from a single input image.
Given an image $I$, the goal is to produce a 3D model $G$ that is consistent with the visible content while inferring geometry and appearance for the occluded or unobserved regions.
Recent image-to-3D generation pipelines~\cite{chen20253dtopia,hunyuan3d2025hunyuan3d_v21,xiang2025structured_trellis} commonly rely on DINO-based visual encoders~\cite{caron2021dino,oquab2023dinov2,simeoni2025dinov3} to extract image conditions for 3D asset generation, which potentially compress spatial details and lead to \textit{detail attenuation}.
Therefore, we propose a feature-enhanced method by blending global and local features for 3D asset generation.

Specifically, existing image-to-3D models~\cite{hunyuan3d2025hunyuan3d_v21,xiang2025trellis2,xiang2025structured_trellis} can be roughly separated into two stages: the shape generation stage and the texture generation stage. 
The first shape stage mainly deals with the structural shape and geometry generations, while the second texture stage generates textures for the shape from the first stage.
Our observed \textit{detail attenuation} problem mainly occurs during the texture generation phase.
Below, we present two observations on TRELLIS.2 that reveal the compositionality and compatibility of DINOv3-encoded features, while these findings are generalizable to other 3D generation models.

\begin{figure}[t]
\centering
\includegraphics[width=0.99\textwidth]{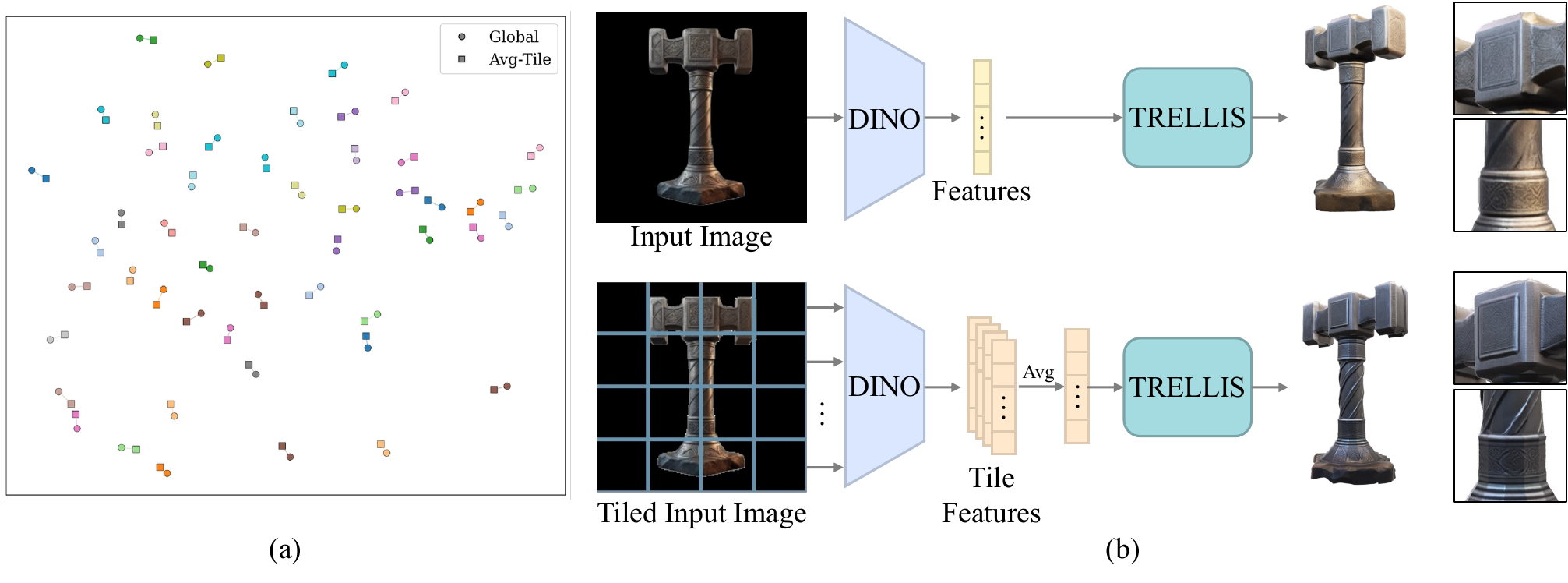}
\vspace{-4mm}
\caption{
Two key observations underpin \ourmethod. 
(a) Compositionality: Global embedding and the average tile embedding tend to cluster in the feature space. Here, the same color denotes the same input image but with global embeddings or \textit{average tile embeddings}.
(b) Compatibility: Average tile embeddings improve fine-grained detail preservation for image-to-3D generation, yet it introduces minor defects in global structural consistency. For instance, the base structure of the generated hammer fails to maintain the regular square shape observed in the baseline output. 
}
\vspace{-4mm}
\label{fig:observations}
\end{figure}

\minisection{Observation 1:  Compositionality of embeddings in the feature space.}
We first investigate whether global image representations can be decomposed into local components.
Specifically, we apply t-SNE to visualize two types of embeddings: the global embedding obtained by directly encoding the input image, the \textit{average tile embedding} derived by encoding \textit{divided image tiles} individually and averaging them.
As shown in~\Cref{fig:observations}, these embeddings tend to cluster closely together in the feature space, suggesting a form of compositionality in the latent space. 

\minisection{Observation 2: Compatibility of embeddings for image-to-3D generation.}
We further examine whether local feature representations can directly replace global conditioning in image-to-3D generation.
Specifically, we input the \textit{average tile embedding}, derived from image tiles to the image-to-3D model for replacing the original global image embeddings.
As shown in~\Cref{fig:observations}, this blended embedding can be directly used for image-to-3D generation, improving fine-grained details while preserving the rough structure.
This suggests that local embeddings can be consumed by the pretrained generator and improves some local details, but direct replacement may disturb global structure.

To summarize, these observations reveal an \textit{image feature additivity} property in the DINO image encoder: global representations can be decomposed into local components (compositionality), and these components can be directly utilized for generation (compatibility). 
We leverage this inherent property to preserve and enhance high-fidelity details during image-to-3D generation.

\subsection{\ourmethod: Blended Tile Conditioning for Image-to-3D Generation}
\label{subsec:overview}

\begin{figure}[t]
    \centering
    \includegraphics[width=0.99\textwidth]{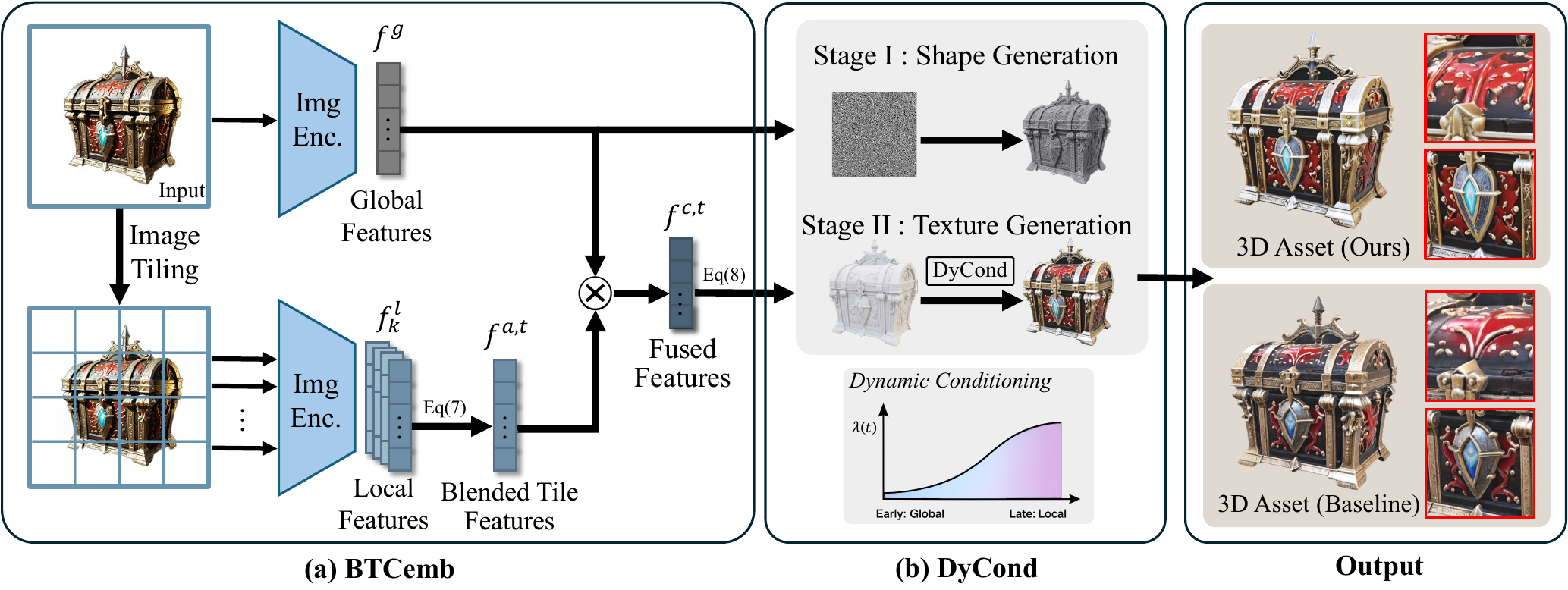}
    \vspace{-3mm}
    \caption{
    The overall pipeline of \ourmethod, which is composed of: (a) the blended tile conditioning embedding (\btcembed) divides the input image into patches to enhance the feature representation; and (b) the dynamic conditioning (\dycond) mechanism dynamically blends global and local features aligning with the coarse-to-fine generation, addressing the detail attenuation problem. 
    }
    \vspace{-4mm}
    \label{fig:method}
\end{figure}

\textbf{Method Overview.}
Our proposed \ourmethod consists of two components:
(1) \textit{Blended Tile Conditioning Embedding} (\btcembed), which extracts local embeddings from overlapping image tiles and aggregates them as local conditions for feature optimization;
and (2) \textit{Dynamic Conditioning} (\dycond), which progressively integrates local conditioning as generation moves from global structure toward fine-grained details.
\ourmethod works during the \textit{texture generation stage} of the image-to-3D models to enhance the detail preservation.
The overall pipeline is illustrated in~\Cref{fig:method}.

\subsubsection{Blended Tile Conditioning Embedding (\btcembed)}
\label{subsubsec:tile}

Existing image-to-3D pipelines~\cite{hunyuan3d2025hunyuan3d_v21,xiang2025trellis2,xiang2025structured_trellis} typically employ an image encoder $E$, such as DINOv3~\cite{simeoni2025dinov3} or DINOv2~\cite{oquab2023dinov2}, to extract a global condition $f^g$ from the input image $I$.
Specifically, the input image is fed into the image encoder to obtain the global feature $f^g$.
To mitigate the loss of spatial details inherent in the global feature, we divide the input image into $K=N \times N$ image tiles, denoted as $\{I_k\}_{k=1}^K$, and encode them individually to obtain the local tile features $f^l_k$:
\begin{gather}
f^g = E(I), \\ 
f^l_k = E(I_k), \quad \{I_k\}_{k=1}^{K} = \mathrm{Image\_Tiling}(I), \quad k=1,\dots,K.
\end{gather}

While local tile features preserve finer visual details than the global feature, not all tiles are equally informative, as background-dominated tiles may provide limited evidence about the target object. For single-object generation, we therefore use foreground coverage to determine each tile's contribution.
Let $M(\mathbf{x})\in[0,1]$ denote the foreground mask obtained from the input image and $\Omega_k$ the pixel region corresponding to tile $I_k$. We define the tile priority score as:
\begin{equation} e_k=\frac{1}{|\Omega_k|}\sum_{\mathbf{x}\in\Omega_k}M(\mathbf{x}),\qquad k=1,\ldots,K. \label{eq:tile_score} \end{equation}

The scores are directly normalized into blending weights, and the local features are aggregated to obtain the Blended Tile Conditioning Embedding (\btcembed):
\begin{equation} w_k=\frac{e_k}{\sum_{j=1}^{K}e_j},\qquad f^a=\sum_{k=1}^{K}w_k f_k^l. \label{eq:btcemb} \end{equation}
This aggregation emphasizes tiles with higher foreground coverage. Alongside the aggregated condition $f^a$, we retain the individual tile features and their priority scores for the subsequent dynamic conditioning process.
Overall, the blended tile embedding \btcembed $f^a$ preserves informative features, complementing the global feature for image-to-3D generation.

\subsubsection{Dynamic Conditioning (\dycond)}
\label{subsubsec:dycond}

\begin{figure}[t]
    \centering
    \includegraphics[width=0.99\textwidth]{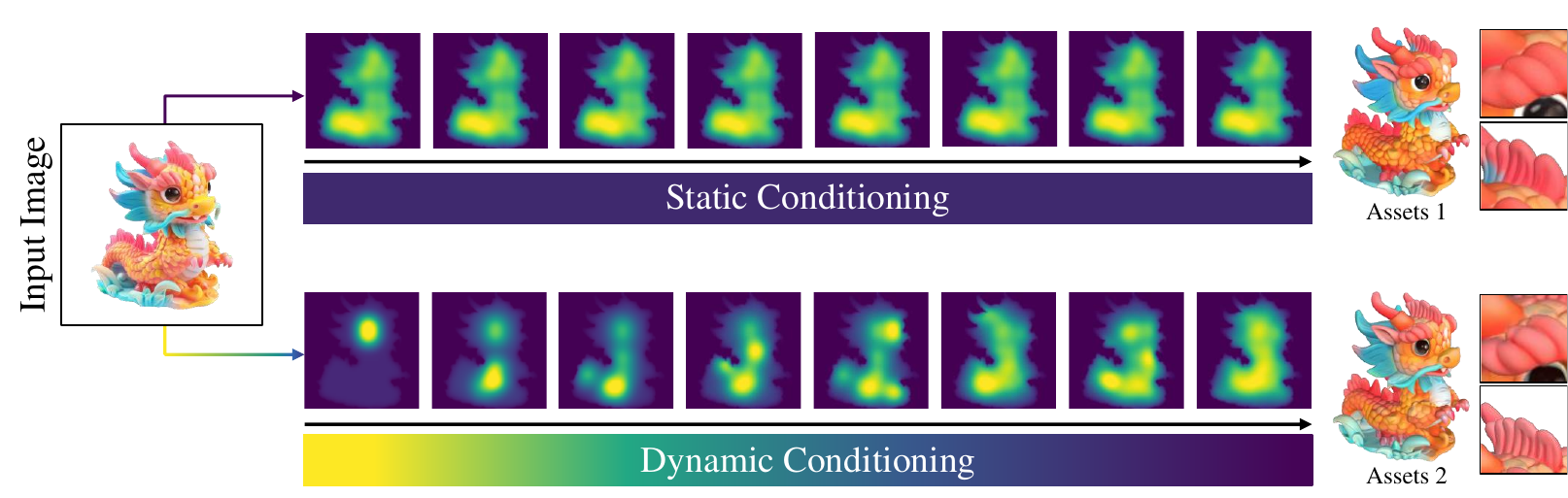}
    \vspace{-2mm}
    \caption{Visualization of the blending weights along time steps and 3D asset generations by two different conditioning mechanisms:\textit{ static conditioning} vs. \textit{dynamic conditioning} (\dycond). Compared with the static conditioning setup, our dynamic conditioning (\dycond) mechanism achieves further refinement of fine-grained texture details.
    }
    \vspace{-3mm}
    \label{fig:progressive_support_expansion}
\end{figure}

Given our proposed \btcembed $f^a$ as the weighted aggregation of local features $f^l$, an intuitive approach would be to \textit{statically fuse} it with the global feature throughout the coarse-to-fine texture generation trajectory as $f^c=
(1-\alpha) \cdot f^g + \alpha \cdot f^a$, which we term \textit{static conditioning}. 
However, this static conditioning embedding $f^c$ consistently focuses on the global object across all generation stages and fails to progressively enhance fine-grained details along the flow-based generation process, as can be observed from the top row in~\Cref{fig:progressive_support_expansion}. 
To mitigate the detail attenuation during 3D asset generation, we introduce a \textit{dynamic conditioning} (\dycond) mechanism that balances detail enhancement and structural consistency throughout the generation pipeline.

Specifically, we progressively integrate the tile features following the coarse-to-fine paradigm of 3D asset generation.
Starting from the static tile weights $\{w_k\}_{k=1}^{K}$ produced by~\cref{eq:btcemb}, we further extend them to timestep-dependent tile weights. 
For \textit{simplicity}, we sort the tiles according to their static weights such that $w_1 \ge w_2 \ge \cdots \ge w_K$.

Let $t \in [0, 1]$ denote the flow matching trajectory, where $t = 0$ and $t = 1$ represent the source distribution and the target, respectively. We use a single slope parameter $\beta \ge 1$ to control the progressive activation of tile conditions:
\begin{equation}
\lambda(t)=\mathrm{clip}(\beta t-\beta+1,\,0,\,1),
\end{equation}
which remains zero in the early stage, starts increasing at $t=1-\frac{1}{\beta}$, and reaches $1$ at $t=1$. Based on $\lambda(t)$, we define the timestep-dependent tile weights as:
\begin{equation}
w_k^t=
\frac{
\operatorname{sigmoid}\!\left(\beta\left(\lambda(t)-\frac{k-1}{K-1}\right)\right)\, w_k
}{
\sum_{j=1}^{K}
\operatorname{sigmoid}\!\left(\beta\left(\lambda(t)-\frac{j-1}{K-1}\right)\right)\, w_j
},
\end{equation}
where $\beta$ jointly controls the slope of the conditioning ramp and the sharpness of progressive tile activation. 
In this way, tiles with larger static weights are emphasized earlier, while lower-weight tiles are gradually introduced as generation proceeds. 
The \textit{dynamic blended tile conditioning embedding} \btcembed by $t$ is then computed as:
\begin{equation}
f^{a,t}=\sum_{k=1}^{K} w_k^t \cdot f_k^l.
\end{equation}
The final \textit{fused conditioning feature} $f^{c,t}$ for image-to-3D generation by time $t$ is defined as:
\begin{equation}
f^{c,t}=
(1-\alpha) \cdot f^g + \alpha \cdot f^{a,t}.
\label{eq:cond_mix}
\end{equation}

This design keeps the global condition dominant in the early high-noise stage and gradually introduces dynamic tile conditions \btcembed in the later low-noise stage, thereby enhancing fine details while preserving overall structural consistency. A schematic illustration of the progressive tile activation process is shown in the bottom of~\Cref{fig:progressive_support_expansion}.

\section{Experiments}
\label{sec:experiments}

\subsection{Evaluation Setups}
\label{subsec:eval_setup}

\minisection{Evaluation Datasets.}
We evaluate BTC3D on 3D-Arena~\cite{3d-arena} and Toys4K~\cite{stojanov2021using_toys4k}. 3D-Arena is a public image-to-3D benchmark with diverse object-centric prompts and reference assets, providing a standardized testbed for comparing representative image-to-3D generation methods under a unified protocol. Toys4K is a 3D object dataset with the largest number of object categories currently available.
More details about benchmarks are provided in~\Cref{appendix:eval_dataset}

\minisection{Comparison Methods.}
We compare \ourmethod with representative image-to-3D baselines, including Hi3DGen~\cite{ye2025hi3dgen}, TripoSG~\cite{li2025triposg}, 3DTopia-XL~\cite{chen20253dtopia},  Hunyuan3D-2.1~\cite{hunyuan3d2025hunyuan3d_v21}, TRELLIS~\cite{xiang2025structured_trellis}, and TRELLIS.2~\cite{xiang2025trellis2}. 
Note that the Hi3DGen and TripoSG are geometry generation methods while the others work on both geometry and texture.
For fair evaluation, all generated 3D assets are rendered into multi-view images at a fixed resolution of $1024\times1024$ using the same rendering protocol, which preserves high-fidelity visual details while avoiding resolution-related evaluation bias.

\minisection{Implementation Details.}
For the image-to-3D models, we select TRELLIS, TRELLIS.2 and Hunyuan3D-2.1 as the backbones to validate the effectiveness of \ourmethod\ across different DINO encoders and diverse 3D latent representation forms.
Following previous papers~\citep{xiang2025trellis2,xiang2025structured_trellis,hunyuan3d2025hunyuan3d_v21}, we use RemBG library to remove the background for condition images.
For hyperparameters, we set $\beta=2, \alpha=0.4$ and $N=3$ to achieve the best trade-offs.
All experiments are conducted on an NVIDIA RTX PRO 6000 GPU. 

\minisection{Evaluation Metrics.}
We evaluate generated 3D assets using complementary metrics. CLIP-I~\cite{radford2021clip} and LPIPS~\cite{zhang2018lpips} measure semantic alignment and perceptual similarity between the input image and rendered views, respectively. PSNR and SSIM~\cite{wang2004ssim} provide auxiliary assessments of pixel-level fidelity and local structural similarity. ULIP-2~\cite{xue2024ulip} and Uni3D~\cite{zhou2023uni3d} assess image-to-asset alignment using native 3D representations. 

\begin{table*}[t]
\caption{\textbf{Quantitative comparison on 3D-Arena and Toys4K benchmarks.} Dashes indicate unavailable results or metrics not applicable to shape-only or untextured outputs. Higher is better for PSNR, SSIM, CLIP-I, ULIP-2, and Uni3D, while lower is better for LPIPS. Best results are shown in \textbf{bold} and second-best results are \underline{underlined}. Note that only our method \ourmethod is \textit{training-free}.}
\vspace{-2mm}
\centering
\scriptsize
\setlength{\tabcolsep}{2.4pt}
\renewcommand{\arraystretch}{1.13}
\resizebox{0.999\textwidth}{!}{
\begin{tabular}{l*{13}{c}}
\toprule
\multirow{2}{*}{Method} & \multirow{2}{*}{\makecell{Train\\Free}} & \multicolumn{6}{c}{3D-Arena} & \multicolumn{6}{c}{Toys4K} \\
\cmidrule(lr){3-8} \cmidrule(lr){9-14}
& & \metric{PSNR}{\uparrow} & \metric{SSIM}{\uparrow} & \metric{CLIP-I}{\uparrow} & \metric{ULIP-2}{\uparrow} & \metric{Uni3D}{\uparrow} & \metric{LPIPS}{\downarrow} & \metric{PSNR}{\uparrow} & \metric{SSIM}{\uparrow} & \metric{CLIP-I}{\uparrow} & \metric{ULIP-2}{\uparrow} & \metric{Uni3D}{\uparrow} & \metric{LPIPS}{\downarrow} \\
\midrule
TripoSG & \xmark & -- & -- & -- & 0.4091 & 0.3738 & -- & -- & -- & -- & 0.4133 & 0.3635 & -- \\
Hi3DGen & \xmark & -- & -- & -- & 0.3924 & 0.3653 & -- & -- & -- & -- & 0.4009 & 0.3607 & -- \\
3DTopia-XL & \xmark & 17.0419 & 0.8134 & 0.5442 & 0.3319 & 0.3104 & 0.4766 & 16.2407 & 0.8218 & 0.5075 & 0.3273 & 0.2720 & 0.7928 \\
\midrule
Hunyuan3D-2.1 & \xmark & 20.6031 & 0.8401 & 0.5497 & 0.3773 & 0.3603 & 0.4886 & 20.6761 & 0.8351 & \underline{0.6437} & 0.4273 & \underline{0.3756} & 0.7805 \\
\rowcolor{gray!12}
\oursmethod & \cmark & \underline{21.1871} & \underline{0.8644} & 0.5568 & 0.3879 & 0.3719 & 0.4789 & \underline{21.0919} & \underline{0.8542} & \textbf{0.6535} & \textbf{0.4320} & \textbf{0.3823} & \underline{0.7767} \\
\midrule
TRELLIS & \xmark & 20.4018 & 0.8356 & 0.5871 & \underline{0.4134} & \underline{0.3765} & 0.4728 & 20.5369 & 0.8262 & 0.6290 & 0.4176 & 0.3637 & 0.7793 \\
\rowcolor{gray!12}
\oursmethod & \cmark & 20.8781 & 0.8589 & \textbf{0.6057} & \textbf{0.4234} & \textbf{0.3914} & \underline{0.4685} & 20.9887 & 0.8534 & 0.6323 & \underline{0.4285} & 0.3747 & \textbf{0.7748} \\
\midrule
TRELLIS.2 & \xmark & 20.6741 & 0.8480 & 0.5802 & 0.3676 & 0.3327 & 0.4693 & 20.7395 & 0.8534 & 0.6295 & 0.4181 & 0.3709 & 0.7841 \\
\rowcolor{gray!12}
\oursmethod & \cmark & \textbf{21.3149} & \textbf{0.8721} & \underline{0.6011} & 0.3837 & 0.3551 & \textbf{0.4597} & \textbf{21.3630} & \textbf{0.8709} & 0.6314 & 0.4243 & 0.3732 & 0.7807 \\
\bottomrule
\end{tabular}
}
\label{tab:main_quantitative_arena_btc3d_assets}
\vspace{-4mm}
\end{table*}

\subsection{Experimental Results}
\minisection{Quantitative Results.}
\Cref{tab:main_quantitative_arena_btc3d_assets} reports the quantitative comparison on the 3D-Arena and Toys4K benchmarks. 
Visual alignment is measured with CLIP-I, perceptual appearance similarity is additionally assessed with LPIPS, pixel-level fidelity and local structural similarity are evaluated with PSNR and SSIM, respectively, and multimodal 3D foundation models including ULIP-2 and Uni3D are employed to evaluate image-to-asset alignment from a 3D-aware representation perspective. 
The results show that our method achieves improvements across all metrics over baselines including the TRELLIS family and Hunyuan3D-2.1. In particular, our gains are more pronounced on native 3D metrics such as ULIP-2 and Uni3D. We also achieve improvements over the baselines on detail-related metrics, including PSNR, SSIM and LPIPS.

\begin{figure}[t]
    \centering
    \vspace{-2mm}
    \includegraphics[width=0.99\textwidth]{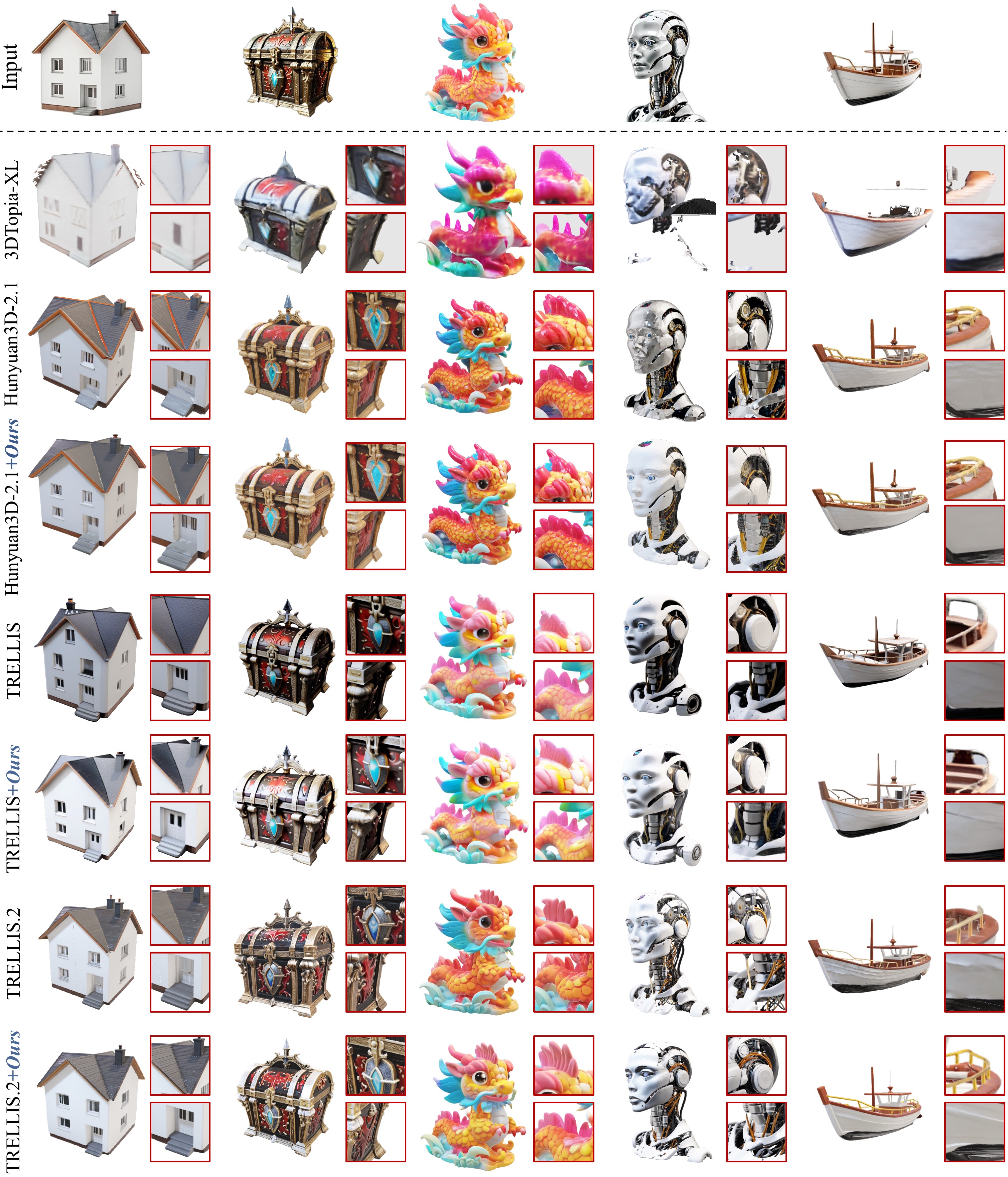}
    \vspace{-3mm}
    \caption{Qualitative comparison of our method \ourmethod with existing image-to-3D approaches. }
    \vspace{-8mm}
    \label{fig:experiments}
\end{figure}

\minisection{Qualitative Results.}
Visual comparisons of our 3D generation quality are presented in~\Cref{fig:experiments}. We exclude TripoSG and Hi3DGen from this comparison, as these methods primarily focus on geometric generation. To demonstrate the effectiveness of \ourmethod, we apply it to three leading open-source image-to-3D models: TRELLIS, TRELLIS.2 and Hunyuan3D-2.1. 
While these backbones can generate visually plausible global geometry from reference inputs, conditioning on a single full-image encoding may not fully account for fine-grained local details. In the illustrated examples, some outputs exhibit smoother surface textures and less distinct local patterns or material cues than those in the input images, reflecting a degree of \textit{detail attenuation}. By incorporating complementary local features, \ourmethod improves the preservation of these appearance details while maintaining coherent global structure, serving as a plug-and-play enhancement to the backbone models.

We further compare the semantic consistency of different generated assets with respect to the input images.
Compared with other methods, the outputs enhanced by \ourmethod better preserve object identity, category-level semantics, and input-specific visual cues.
Although TripoSG and Hi3DGen can produce reasonable geometric structures, their outputs are primarily shape-oriented and lack faithful surface appearance.
Similarly, 3DTopia-XL often captures the coarse object shape but loses fine local correspondences between the generated asset and the input image.
By comparison, \ourmethod maintains semantic alignment while recovering fine-grained textures, sharp material boundaries, and realistic surface appearances.
We also note that existing methods~\cite{chen20253dtopia,hunyuan3d2025hunyuan3d_v21} generally achieve high-fidelity image-to-3D generation through large-scale training regimens. In contrast, \ourmethod enables the restoration of fine-grained visual detail and texture fidelity directly from pre-trained off-the-shelf 3D generation models in a completely \textit{training-free} manner.

\subsection{Ablation Studies}
We perform ablation experiments on the 3D-Arena~\cite{3d-arena} benchmark, adopting TRELLIS.2 as the backbone. The goal is to quantitatively validate the individual contribution of each design.
Additional ablation results are provided in Appendix~\ref{appendix:appendix_ablation}

\begin{figure*}[t]
\centering
\includegraphics[width=0.999\textwidth]{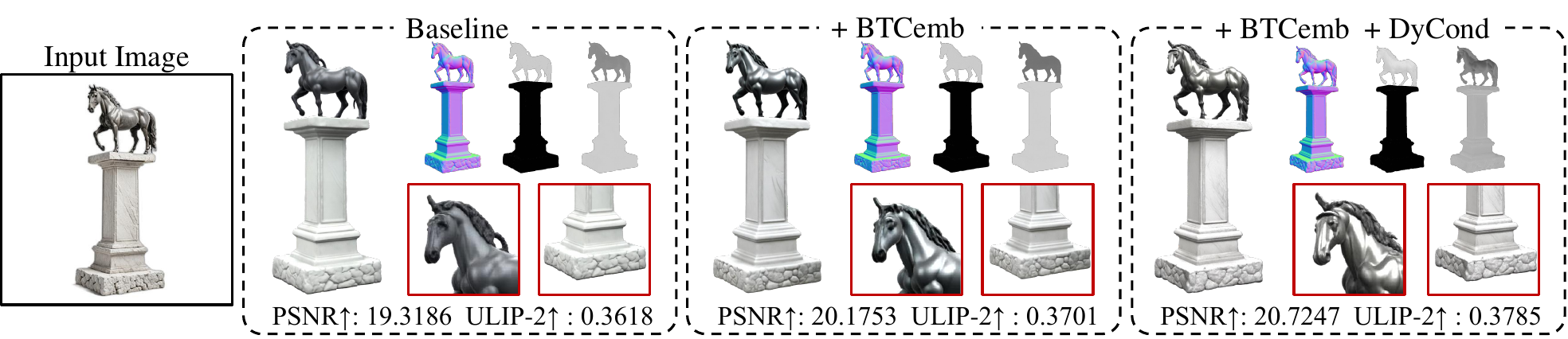}
\vspace{-6mm}
\caption{\textbf{Qualitative ablation of \ourmethod.} From left to right: input image, baseline, baseline with \btcembed, and \ourmethod composing of \btcembed and \dycond. Red boxes highlight generation details.}
\vspace{-1mm}
\label{fig:qualitative_ablation}
\end{figure*}   

\minisection{Ablate each component design.}
First, we evaluate the respective efficacy of our two introduced modules. As shown in~\Cref{tab:ablation_btc3d} and~\Cref{fig:qualitative_ablation}, equipping the TRELLIS.2 baseline with \btcembed\ yields consistent performance gains on both ULIP-2 and Uni3D metrics. This indicates that our foreground-based tile weighting delivers more expressive local cues than vanilla global conditioning alone. Further integrating \dycond\ upon \btcembed\ brings additional performance improvements, demonstrating that progressively injecting tile-level conditioning in the low-noise generation stage is more effective than static conditioning. Combining both modules achieves the best overall results, which verifies the complementary effect of our proposed modules.

\minisection{Ablate \btcembed hyperparameter $N$.}
We ablate the hyperparameter $N$ for tile conditioning in \btcembed, as summarized in~\Cref{tab:ablation_btcemb}. Here, $N$ controls how the input image is divided into tiles. Empirically, $N=3$ achieves the best performance on both ULIP-2 and Uni3D.

\minisection{Ablate \dycond hyperparameter $\beta$.}
We also ablate the hyperparameter $\beta$ for tile conditioning in \dycond, as summarized in~\Cref{tab:ablation_dycond}. In our dynamic conditioning framework, $\beta$ controls the slope of the progressive tile activation schedule. Empirically, $\beta=2$ produces the best results, and all dynamic conditioning configurations with different $\beta$ values consistently outperform the static conditioning.

\minisection{Ablate fusion ratio $\alpha$.}
We ablate the fusion ratio $\alpha$ in \dycond in~\Cref{tab:ablation_fuse}. As the blending coefficient between $f^g$ and $f^{a,t}$, $\alpha$ controls the balance between global coherence and detail enhancement. Empirically, $\alpha=0.4$ achieves the best performance on both ULIP-2 and Uni3D. Smaller $\alpha$ weakens tile conditioning, while larger $\alpha$ slightly harms structural consistency.

\begin{table}[!t]
\centering
\setlength{\tabcolsep}{4pt}
\begin{minipage}[t]{0.281\textwidth}
\centering
\captionof{table}{Ablate designs.}
\vspace{-2mm}
\resizebox{\linewidth}{!}{
\begin{tabular}{lcc}
\toprule
 & ULIP2 $\uparrow$ & Uni3D $\uparrow$ \\
\midrule
Baseline & 0.3676 & 0.3327 \\
+\btcembed & 0.3733 & 0.3446 \\
\makecell[l]{+\dycond\\(full \ourmethod)} & \textbf{0.3837} & \textbf{0.3551} \\
\bottomrule
\end{tabular}
}
\label{tab:ablation_btc3d}
\end{minipage}
\hfill
\begin{minipage}[t]{0.23\textwidth}
\centering
\captionof{table}{Ablate $N$.}
\vspace{-2mm}
\resizebox{0.94\linewidth}{!}{
\begin{tabular}{lcc}
\toprule
 & ULIP2 $\uparrow$ & Uni3D $\uparrow$ \\
\midrule
$N$=2 & 0.3787 & 0.3469 \\
\textbf{$N$=3} & \textbf{0.3837} & \textbf{0.3551} \\
$N$=4 & 0.3805 & 0.3490 \\
$N$=5 & 0.3732 & 0.3378 \\
\bottomrule
\end{tabular}
}
\label{tab:ablation_btcemb}
\end{minipage}
\hfill
\begin{minipage}[t]{0.225\textwidth}
\centering
\captionof{table}{Ablation of $\beta$.}
\vspace{-2mm}
\resizebox{0.98\linewidth}{!}{
\begin{tabular}{lcc}
\toprule
 & ULIP2 $\uparrow$ & Uni3D $\uparrow$ \\
\midrule
Static & 0.3733 & 0.3446 \\
$\beta$=1 & 0.3795 & 0.3488 \\
\textbf{$\beta$=2} & \textbf{0.3837} & \textbf{0.3551} \\
$\beta$=3 & 0.3801 & 0.3529 \\
\bottomrule
\end{tabular}
}
\label{tab:ablation_dycond}
\end{minipage}
\hfill
\begin{minipage}[t]{0.24\textwidth}
\centering
\captionof{table}{Ablation of $\alpha$.}
\vspace{-2mm}
\resizebox{\linewidth}{!}{
\begin{tabular}{lcc}
\toprule
 & ULIP2 $\uparrow$ & Uni3D $\uparrow$ \\
\midrule
Baseline & 0.3676 & 0.3327 \\
$\alpha$=0.3 & 0.3801 & 0.3515 \\
\textbf{$\alpha$=0.4} & \textbf{0.3837} & \textbf{0.3551} \\
$\alpha$=0.5 & 0.3799 & 0.3492 \\
\bottomrule
\end{tabular}
}
\label{tab:ablation_fuse}
\end{minipage}
\vspace{-3mm}
\end{table}

\section{Conclusion}
In this work, we present a \textit{training-free} refinement framework, named \textbf{B}lended \textbf{T}ile \textbf{C}onditioning for Image-to-\textbf{3D} synthesis (\ourmethod).
Our approach \ourmethod starts from our observation about the image feature additivity in image-to-3D generative models. Based on such property, we introduce the blended tile-conditioned embedding (\btcembed) that extracts local conditioning signals from divided image patches, allowing the diffusion model to better preserve fine-grained visual details. 
To stabilize the integration of global and local guidance, we propose a \textit{dynamic conditioning} (\dycond) schedule that gradually increases the influence of tile-level conditioning during later low-noise diffusion regimes, such an operation further enhances the texture quality.
Our method \ourmethod operates entirely at inference time and can be seamlessly integrated into diverse image-to-3D diffusion pipelines. Experimental results demonstrate that the proposed approach improves texture sharpness and visual fidelity while maintaining global structural consistency.

\clearpage

\subsection*{AI use statement}
\label{sec:ai_use_statement}
We used generative AI tools to assist with literature searches, suggest revisions to manuscript text, and check citation formatting and LaTeX layout. We did not use generative AI to develop the research method, implement experimental code, design experiments, generate or process datasets, or interpret experimental results. The authors reviewed the AI-assisted revisions and checked
the suggested references against the original publications. The authors take responsibility for the accuracy, originality, and integrity of the final manuscript.

\bibliographystyle{iclr2027_conference}
\bibliography{longstrings,mybib}

@string{cvpr   = "Proceedings of the IEEE Conference on Computer Vision and Pattern Recognition"}

@string{neurips = "Advances in Neural Information Processing Systems"}

@string{pami =  "{IEEE} Transactions on Pattern Analysis and Machine Intelligence"}

@string{siggraph = "Proceedings of the ACM SIGGRAPH Conference on Computer Graphics"}

@string{ICLR = "International Conference on Learning Representations"}

@string{jun = "June"}

@article{rosenholtz2007visual,
  title={Measuring visual clutter},
  author={Rosenholtz, Ruth and Li, Yuanzhen and Nakano, Lisa},
  journal={Journal of Vision},
  volume={7},
  number={2},
  pages={17--17},
  year={2007}
}

@inproceedings{pech2000diatom,
  title={Diatom autofocusing in brightfield microscopy: a comparative study},
  author={Pech-Pacheco, Jos{\'e} Luis and Crist{\'o}bal, Gabriel and Chamorro-Mart{\'i}nez, Jes{\'u}s and Fern{\'a}ndez-Valdivia, Joaqu{\'i}n},
  booktitle={Proceedings 15th International Conference on Pattern Recognition},
  volume={3},
  pages={314--317},
  year={2000},
  organization={IEEE}
}

@article{pertuz2013analysis,
  title={Analysis of focus measure operators for shape-from-focus},
  author={Pertuz, Said and Puig, Domenec and Garcia, Miguel Angel},
  journal={Pattern Recognition},
  volume={46},
  number={5},
  pages={1415--1432},
  year={2013},
  publisher={Elsevier}
}

@InProceedings{Rombach_2022_CVPR_stablediffusion,
    author    = {Rombach, Robin and Blattmann, Andreas and Lorenz, Dominik and Esser, Patrick and Ommer, Bj\"orn},
    title     = {High-Resolution Image Synthesis With Latent Diffusion Models},
    booktitle = {Proceedings of the IEEE/CVF Conference on Computer Vision and Pattern Recognition (CVPR)},
    month     = {06},
    year      = {2022},
    pages     = {10684-10695}
}

@article{ramesh2022dalle2,
  title={Hierarchical text-conditional image generation with clip latents},
  author={Ramesh, Aditya and Dhariwal, Prafulla and Nichol, Alex and Chu, Casey and Chen, Mark},
  journal={arXiv preprint arXiv:2204.06125},
  year={2022}
}

@inproceedings{radford2021clip,
  title={Learning transferable visual models from natural language supervision},
  author={Radford, Alec and Kim, Jong Wook and Hallacy, Chris and Ramesh, Aditya and Goh, Gabriel and Agarwal, Sandhini and Sastry, Girish and Askell, Amanda and Mishkin, Pamela and Clark, Jack and others},
  booktitle={International conference on machine learning},
  pages={8748--8763},
  year={2021},
  organization={PMLR}
}

@article{wang2023prolificdreamer,
  title={Prolificdreamer: High-fidelity and diverse text-to-3d generation with variational score distillation},
  author={Wang, Zhengyi and Lu, Cheng and Wang, Yikai and Bao, Fan and Li, Chongxuan and Su, Hang and Zhu, Jun},
  journal={Advances in Neural Information Processing Systems},
  volume={36},
  pages={8406--8441},
  year={2023}
}

@article{ho2020ddpm,
  title={Denoising diffusion probabilistic models},
  author={Ho, Jonathan and Jain, Ajay and Abbeel, Pieter},
  journal={Advances in Neural Information Processing Systems},
  volume={33},
  pages={6840--6851},
  year={2020}
}

@inproceedings{caron2021dino,
  title={Emerging properties in self-supervised vision transformers},
  author={Caron, Mathilde and Touvron, Hugo and Misra, Ishan and J{\'e}gou, Herv{\'e} and Mairal, Julien and Bojanowski, Piotr and Joulin, Armand},
  booktitle={Proceedings of the IEEE/CVF international conference on computer vision},
  pages={9650--9660},
  year={2021}
}

@inproceedings{zhang2018lpips,
  title={The unreasonable effectiveness of deep features as a perceptual metric},
  author={Zhang, Richard and Isola, Phillip and Efros, Alexei A and Shechtman, Eli and Wang, Oliver},
  booktitle={Proceedings of the IEEE conference on computer vision and pattern recognition},
  pages={586--595},
  year={2018}
}

@inproceedings{khachatryan2023text2video,
  title={Text2video-zero: Text-to-image diffusion models are zero-shot video generators},
  author={Khachatryan, Levon and Movsisyan, Andranik and Tadevosyan, Vahram and Henschel, Roberto and Wang, Zhangyang and Navasardyan, Shant and Shi, Humphrey},
  booktitle={Proceedings of the IEEE/CVF International Conference on Computer Vision},
  pages={15954--15964},
  year={2023}
}

@misc{oquab2023dinov2,
  title={DINOv2: Learning Robust Visual Features without Supervision},
  author={Oquab, Maxime and Darcet, Timothée and Moutakanni, Theo and Vo, Huy V. and Szafraniec, Marc and Khalidov, Vasil and Fernandez, Pierre and Haziza, Daniel and Massa, Francisco and El-Nouby, Alaaeldin and Howes, Russell and Huang, Po-Yao and Xu, Hu and Sharma, Vasu and Li, Shang-Wen and Galuba, Wojciech and Rabbat, Mike and Assran, Mido and Ballas, Nicolas and Synnaeve, Gabriel and Misra, Ishan and Jegou, Herve and Mairal, Julien and Labatut, Patrick and Joulin, Armand and Bojanowski, Piotr},
  journal={arXiv:2304.07193},
  year={2023}
}

@article{poole2023dreamfusion,
  author = {Poole, Ben and Jain, Ajay and Barron, Jonathan T. and Mildenhall, Ben},
  title = {{DreamFusion: Text-to-3D using 2D Diffusion}},
  journal = ICLR,
  year = {2023},
}

@article{podell2023sdxl,
  title={Sdxl: Improving latent diffusion models for high-resolution image synthesis},
  author={Podell, Dustin and English, Zion and Lacey, Kyle and Blattmann, Andreas and Dockhorn, Tim and M{\"u}ller, Jonas and Penna, Joe and Rombach, Robin},
  journal=iclr,
  year={2024}
}

@misc{jayasumana2024rethinking_fid,
      title={Rethinking FID: Towards a Better Evaluation Metric for Image Generation}, 
      author={Sadeep Jayasumana and Srikumar Ramalingam and Andreas Veit and Daniel Glasner and Ayan Chakrabarti and Sanjiv Kumar},
      year={2024},
      eprint={2401.09603},
      archivePrefix={arXiv},
      primaryClass={cs.CV}
}

@article{wu2025direct3d,
  title={Direct3d-s2: Gigascale 3d generation made easy with spatial sparse attention},
  author={Wu, Shuang and Lin, Youtian and Zhang, Feihu and Zeng, Yifei and Yang, Yikang and Bao, Yajie and Qian, Jiachen and Zhu, Siyu and Cao, Xun and Torr, Philip and others},
  journal=neurips,
  year={2025}
}

@misc{3d-arena,
      title={3D Arena: An Open Platform for Generative 3D Evaluation}, 
      author={Dylan Ebert},
      year={2025},
      eprint={2506.18787},
      archivePrefix={arXiv},
      primaryClass={cs.CV},
      url={https://arxiv.org/abs/2506.18787}, 
}

@article{simeoni2025dinov3,
  title={Dinov3},
  author={Sim{\'e}oni, Oriane and Vo, Huy V and Seitzer, Maximilian and Baldassarre, Federico and Oquab, Maxime and Jose, Cijo and Khalidov, Vasil and Szafraniec, Marc and Yi, Seungeun and Ramamonjisoa, Micha{\"e}l and others},
  journal={arXiv preprint arXiv:2508.10104},
  year={2025}
}

@inproceedings{wang2024luciddreaming,
  title={Luciddreaming: Controllable object-centric 3d generation},
  author={Wang, Zhaoning and Li, Ming and Chen, Chen},
  booktitle={European Conference on Computer Vision},
  pages={304--320},
  year={2025},
  organization={Springer}
}

@inproceedings{tang2023dreamgaussian,
 author = {Tang, Jiaxiang and Ren, Jiawei and Zhou, Hang and Liu, Ziwei and Zeng, Gang},
 booktitle = {International Conference on Learning Representations},
 pages = {33879--33896},
 title = {DreamGaussian: Generative Gaussian Splatting for Efficient 3D Content Creation},
 volume = {2024},
 year = {2024}
}

@article{li2025step1x_3d,
  title={Step1x-3d: Towards high-fidelity and controllable generation of textured 3d assets},
  author={Li, Weiyu and Zhang, Xuanyang and Sun, Zheng and Qi, Di and Li, Hao and Cheng, Wei and Cai, Weiwei and Wu, Shihao and Liu, Jiarui and Wang, Zihao and others},
  journal={arXiv preprint arXiv:2505.07747},
  year={2025}
}

@inproceedings{ye2025hi3dgen,
  title={Hi3dgen: High-fidelity 3d geometry generation from images via normal bridging},
  author={Ye, Chongjie and Wu, Yushuang and Lu, Ziteng and Chang, Jiahao and Guo, Xiaoyang and Zhou, Jiaqing and Zhao, Hao and Han, Xiaoguang},
  booktitle={Proceedings of the IEEE/CVF International Conference on Computer Vision},
  pages={25050--25061},
  year={2025}
}

@inproceedings{stojanov2021using_toys4k,
  title={Using shape to categorize: Low-shot learning with an explicit shape bias},
  author={Stojanov, Stefan and Thai, Anh and Rehg, James M},
  booktitle={Proceedings of the IEEE/CVF conference on computer vision and pattern recognition},
  pages={1798--1808},
  year={2021}
}

@article{hu2024token_merging_tome,
  title={Token Merging for Training-Free Semantic Binding in Text-to-Image Synthesis},
  author={Hu, Taihang and Li, Linxuan and van de Weijer, Joost and Gao, Hongcheng and Shahbaz Khan, Fahad and Yang, Jian and Cheng, Ming-Ming and Wang, Kai and Wang, Yaxing},
  journal={Advances in Neural Information Processing Systems},
  volume={37},
  pages={137646--137672},
  year={2024}
}

@inproceedings{xiang2025structured_trellis,
  title={Structured 3d latents for scalable and versatile 3d generation},
  author={Xiang, Jianfeng and Lv, Zelong and Xu, Sicheng and Deng, Yu and Wang, Ruicheng and Zhang, Bowen and Chen, Dong and Tong, Xin and Yang, Jiaolong},
  booktitle={Proceedings of the Computer Vision and Pattern Recognition Conference},
  pages={21469--21480},
  year={2025}
}

@misc{hunyuan3d2025hunyuan3d_v21,
    title={Hunyuan3D 2.1: From Images to High-Fidelity 3D Assets with Production-Ready PBR Material},
    author={Tencent Hunyuan3D Team},
    year={2025},
    eprint={2506.15442},
    archivePrefix={arXiv},
    primaryClass={cs.CV}
}

@misc{hunyuan3d22025tencent_v2,
    title={Hunyuan3D 2.0: Scaling Diffusion Models for High Resolution Textured 3D Assets Generation},
    author={Tencent Hunyuan3D Team},
    year={2025},
    eprint={2501.12202},
    archivePrefix={arXiv},
    primaryClass={cs.CV}
}

@misc{yang2024hunyuan3d_v1,
    title={Hunyuan3D 1.0: A Unified Framework for Text-to-3D and Image-to-3D Generation},
    author={Tencent Hunyuan3D Team},
    year={2024},
    eprint={2411.02293},
    archivePrefix={arXiv},
    primaryClass={cs.CV}
}

@article{
    xiang2025trellis2,
    title={Native and Compact Structured Latents for 3D Generation},
    author={Xiang, Jianfeng and Chen, Xiaoxue and Xu, Sicheng and Wang, Ruicheng and Lv, Zelong and Deng, Yu and Zhu, Hongyuan and Dong, Yue and Zhao, Hao and Yuan, Nicholas Jing and Yang, Jiaolong},
    journal={Tech report},
    year={2025}
}

@article{lipman2022flow_matching,
  title={Flow matching for generative modeling},
  author={Lipman, Yaron and Chen, Ricky TQ and Ben-Hamu, Heli and Nickel, Maximilian and Le, Matt},
  journal=iclr,
  year={2023}
}

@article{liu2022_rectified_flow,
  title={Flow straight and fast: Learning to generate and transfer data with rectified flow},
  author={Liu, Xingchao and Gong, Chengyue and Liu, Qiang},
  journal=iclr,
  year={2023}
}

@inproceedings{qin2025free_sadis,
 author = {Qin, Jiang and Gomez-Villa, Alexandra and Li, Senmao and Yang, Shiqi and Wang, Yaxing and Wang, Kai and van de Weijer, Joost},
 booktitle = {Advances in Neural Information Processing Systems},
 doi = {10.52202/085713-4487},
 editor = {D. Belgrave and C. Zhang and H. Lin and R. Pascanu and P. Koniusz and M. Ghassemi and N. Chen},
 pages = {134488--134522},
 publisher = {Curran Associates, Inc.},
 title = {Free-Lunch Color-Texture Disentanglement for Stylized Image Generation},
 volume = {38, Main Conference},
 year = {2025}
}

@inproceedings{shi2023mvdream,
  title={Mvdream: Multi-view diffusion for 3d generation},
  author={Shi, Yichun and Wang, Peng and Ye, Jianglong and Mai, Long and Li, Kejie and Yang, Xiao},
  booktitle={International conference on learning representations},
  volume={2024},
  pages={39838--39859},
  year={2024}
}

@article{yu2024viewcrafter,
  title={Viewcrafter: Taming video diffusion models for high-fidelity novel view synthesis},
  author={Yu, Wangbo and Xing, Jinbo and Yuan, Li and Hu, Wenbo and Li, Xiaoyu and Huang, Zhipeng and Gao, Xiangjun and Wong, Tien-Tsin and Shan, Ying and Tian, Yonghong},
  journal=pami,
  year={2025}
}

@article{shriram2024realmdreamer,
  title={Realmdreamer: Text-driven 3d scene generation with inpainting and depth diffusion},
  author={Shriram, Jaidev and Trevithick, Alex and Liu, Lingjie and Ramamoorthi, Ravi},
  journal={3DV 2025},
  year={2025}
}

@Article{kerbl2023_3dgs,
  author       = {Kerbl, Bernhard and Kopanas, Georgios and Leimk{\"u}hler, Thomas and Drettakis, George},
  title        = {3D Gaussian Splatting for Real-Time Radiance Field Rendering},
  journal      = {ACM Transactions on Graphics},
  number       = {4},
  volume       = {42},
  month        = {July},
  year         = {2023}
}

@inproceedings{vahdat2022lion,
 author = {Zeng, Xiaohui and Vahdat, Arash and Williams, Francis and Gojcic, Zan and Litany, Or and Fidler, Sanja and Kreis, Karsten},
 booktitle = {Advances in Neural Information Processing Systems},
 doi = {10.52202/068431-0728},
 editor = {S. Koyejo and S. Mohamed and A. Agarwal and D. Belgrave and K. Cho and A. Oh},
 pages = {10021--10039},
 publisher = {Curran Associates, Inc.},
 title = {LION: Latent Point Diffusion Models for 3D Shape Generation},
 volume = {35},
 year = {2022}
}

@inproceedings{ren2024xcube,
  title={Xcube: Large-scale 3d generative modeling using sparse voxel hierarchies},
  author={Ren, Xuanchi and Huang, Jiahui and Zeng, Xiaohui and Museth, Ken and Fidler, Sanja and Williams, Francis},
  booktitle={Proceedings of the IEEE/CVF conference on computer vision and pattern recognition},
  pages={4209--4219},
  year={2024}
}

@inproceedings{zeng2024paint3d,
  title={Paint3d: Paint anything 3d with lighting-less texture diffusion models},
  author={Zeng, Xianfang and Chen, Xin and Qi, Zhongqi and Liu, Wen and Zhao, Zibo and Wang, Zhibin and Fu, Bin and Liu, Yong and Yu, Gang},
  booktitle={Proceedings of the IEEE/CVF conference on computer vision and pattern recognition},
  pages={4252--4262},
  year={2024}
}

@inproceedings{cheng2023sdfusion,
  title={Sdfusion: Multimodal 3d shape completion, reconstruction, and generation},
  author={Cheng, Yen-Chi and Lee, Hsin-Ying and Tulyakov, Sergey and Schwing, Alexander G and Gui, Liang-Yan},
  booktitle={Proceedings of the IEEE/CVF conference on computer vision and pattern recognition},
  pages={4456--4465},
  year={2023}
}

@inproceedings{deitke2023objaverse,
  title={Objaverse: A universe of annotated 3d objects},
  author={Deitke, Matt and Schwenk, Dustin and Salvador, Jordi and Weihs, Luca and Michel, Oscar and VanderBilt, Eli and Schmidt, Ludwig and Ehsani, Kiana and Kembhavi, Aniruddha and Farhadi, Ali},
  booktitle={Proceedings of the IEEE/CVF conference on computer vision and pattern recognition},
  pages={13142--13153},
  year={2023}
}

@article{deitke2023objaverse_plus,
  title={Objaverse-xl: A universe of 10m+ 3d objects},
  author={Deitke, Matt and Liu, Ruoshi and Wallingford, Matthew and Ngo, Huong and Michel, Oscar and Kusupati, Aditya and Fan, Alan and Laforte, Christian and Voleti, Vikram and Gadre, Samir Yitzhak and others},
  journal={Advances in Neural Information Processing Systems},
  volume={36},
  pages={35799--35813},
  year={2023}
}

@inproceedings{chen20253dtopia,
  title={3dtopia-xl: Scaling high-quality 3d asset generation via primitive diffusion},
  author={Chen, Zhaoxi and Tang, Jiaxiang and Dong, Yuhao and Cao, Ziang and Hong, Fangzhou and Lan, Yushi and Wang, Tengfei and Xie, Haozhe and Wu, Tong and Saito, Shunsuke and others},
  booktitle={Proceedings of the Computer Vision and Pattern Recognition Conference},
  pages={26576--26586},
  year={2025}
}

@article{li2025triposg,
  title={Triposg: High-fidelity 3d shape synthesis using large-scale rectified flow models},
  author={Li, Yangguang and Zou, Zi-Xin and Liu, Zexiang and Wang, Dehu and Liang, Yuan and Yu, Zhipeng and Liu, Xingchao and Guo, Yuan-Chen and Liang, Ding and Ouyang, Wanli and others},
  journal={IEEE Transactions on Pattern Analysis and Machine Intelligence},
  year={2025},
  publisher={IEEE}
}

@article{hong2023lrm,
  title={Lrm: Large reconstruction model for single image to 3d},
  author={Hong, Yicong and Zhang, Kai and Gu, Jiuxiang and Bi, Sai and Zhou, Yang and Liu, Difan and Liu, Feng and Sunkavalli, Kalyan and Bui, Trung and Tan, Hao},
  journal=iclr,
  year={2024}
}

@inproceedings{du2025hierarchical,
  title={Hierarchical neural semantic representation for 3d semantic correspondence},
  author={Du, Keyu and Hu, Jingyu and Li, Haipeng and Xu, Hao and Huang, Haibin and Fu, Chi-Wing and Liu, Shuaicheng},
  booktitle={Proceedings of the SIGGRAPH Asia 2025 Conference Papers},
  pages={1--11},
  year={2025}
}

@article{hu2024neural,
  title={Neural wavelet-domain diffusion for 3d shape generation, inversion, and manipulation},
  author={Hu, Jingyu and Hui, Ka-Hei and Liu, Zhengzhe and Li, Ruihui and Fu, Chi-Wing},
  journal={ACM transactions on graphics},
  volume={43},
  number={2},
  pages={1--18},
  year={2024},
  publisher={ACM New York, NY, USA}
}

@inproceedings{girdhar2023imagebind,
  title={Imagebind: One embedding space to bind them all},
  author={Girdhar, Rohit and El-Nouby, Alaaeldin and Liu, Zhuang and Singh, Mannat and Alwala, Kalyan Vasudev and Joulin, Armand and Misra, Ishan},
  booktitle={Proceedings of the IEEE/CVF conference on computer vision and pattern recognition},
  pages={15180--15190},
  year={2023}
}

@article{mikolov2013efficient_word2vec,
  title={Efficient estimation of word representations in vector space},
  author={Mikolov, Tomas and Chen, Kai and Corrado, Greg and Dean, Jeffrey},
  journal={arXiv preprint arXiv:1301.3781},
  year={2013}
}

@inproceedings{xue2024ulip,
  title={Ulip-2: Towards scalable multimodal pre-training for 3d understanding},
  author={Xue, Le and Yu, Ning and Zhang, Shu and Panagopoulou, Artemis and Li, Junnan and Mart{\'\i}n-Mart{\'\i}n, Roberto and Wu, Jiajun and Xiong, Caiming and Xu, Ran and Niebles, Juan Carlos and others},
  booktitle={Proceedings of the IEEE/CVF Conference on Computer Vision and Pattern Recognition},
  pages={27091--27101},
  year={2024}
}

@article{zhou2023uni3d,
  title={Uni3d: Exploring unified 3d representation at scale},
  author={Zhou, Junsheng and Wang, Jinsheng and Ma, Baorui and Liu, Yu-Shen and Huang, Tiejun and Wang, Xinlong},
  journal=iclr,
  year={2024}
}

@INPROCEEDINGS{li2025voxhammer,
  author={Li, Lin and Huang, Zehuan and Feng, Haoran and Zhuang, Gengxiong and Chen, Rui and Guo, Chunchao and Sheng, Lu},
  booktitle={2026 International Conference on 3D Vision (3DV)}, 
  title={VoxHammer: Training-Free Precise and Coherent 3D Editing in Native 3D Space}, 
  year={2026},
  volume={},
  number={},
  pages={1281-1292}
}

@INPROCEEDINGS{Qiu2024richdreamer,
  author={Qiu, Lingteng and Chen, Guanying and Gu, Xiaodong and Zuo, Qi and Xu, Mutian and Wu, Yushuang and Yuan, Weihao and Dong, Zilong and Bo, Liefeng and Han, Xiaoguang},
  booktitle={2024 IEEE/CVF Conference on Computer Vision and Pattern Recognition (CVPR)}, 
  title={RichDreamer: A Generalizable Normal-Depth Diffusion Model for Detail Richness in Text-to-3D}, 
  year={2024},
  volume={},
  number={},
  pages={9914-9925}
}

@article{liu2024one2345,                                                      
    title={One-2-3-45++: Fast Single Image to 3D Objects with Consistent        
Multi-View Generation and 3D Diffusion},                                        
    author={Liu, Minghua and others},                                           
    journal={arXiv preprint arXiv:2311.07885},                                  
    year={2023}                                                                 
  }

@article{khader2023denoising_3dmedical,
  title={Denoising diffusion probabilistic models for 3D medical image generation},
  author={Khader, Firas and M{\"u}ller-Franzes, Gustav and Tayebi Arasteh, Soroosh and Han, Tianyu and Haarburger, Christoph and Schulze-Hagen, Maximilian and Schad, Philipp and Engelhardt, Sandy and Bae{\ss}ler, Bettina and Foersch, Sebastian and others},
  journal={Scientific reports},
  volume={13},
  number={1},
  pages={7303},
  year={2023},
  publisher={Nature Publishing Group UK London}
}

@article{ke20243d_robot,
  title={3d diffuser actor: Policy diffusion with 3d scene representations},
  author={Ke, Tsung-Wei and Gkanatsios, Nikolaos and Fragkiadaki, Katerina},
  journal={arXiv preprint arXiv:2402.10885},
  year={2024}
}

@article{xu2024sketch2scene_3dgame,
  title={Sketch2Scene: Automatic Generation of Interactive 3D Game Scenes from User's Casual Sketches},
  author={Xu, Yongzhi and Ng, Yonhon and Wang, Yifu and Sa, Inkyu and Duan, Yunfei and Sun, Zhenhong and Li, Yang and Ji, Pan and Li, Hongdong},
  journal={arXiv preprint arXiv:2408.04567},
  year={2024}
}

@article{zhang2025_3dgenbench,
  title        = {3DGen-Bench: Comprehensive Benchmark Suite for 3D Generative Models},
  author       = {Zhang, Yuhan and Zhang, Mengchen and Wu, Tong and Wang, Tengfei and Wetzstein, Gordon and Lin, Dahua and Liu, Ziwei},
  journal      = {arXiv preprint arXiv:2503.21745},
  year         = {2025},
  eprint       = {2503.21745},
  archivePrefix= {arXiv},
  primaryClass = {cs.CV}
}

@article{wang2004ssim,
  title   = {Image Quality Assessment: From Error Visibility to Structural Similarity},
  author  = {Wang, Zhou and Bovik, Alan C. and Sheikh, Hamid R. and Simoncelli, Eero P.},
  journal = {IEEE Transactions on Image Processing},
  volume  = {13},
  number  = {4},
  pages   = {600--612},
  year    = {2004},
  doi     = {10.1109/TIP.2003.819861}
}

@article{feng2025seed3d,
  title         = {Seed3D 1.0: From Images to High-Fidelity Simulation-Ready 3D Assets},
  author        = {Feng, Jiashi and Li, Xiu and Lin, Jing and Liu, Jiahang and Liu, Gaohong and Lou, Weiqiang and Ma, Su and Shi, Guang and Wang, Qinlong and Wang, Jun and Xu, Zhongcong and Yi, Xuanyu and Yu, Zihao and Zhang, Jianfeng and Zhu, Yifan and Chen, Rui and Chi, Jinxin and Du, Zixian and Han, Li and Huang, Lixin and Jiang, Kaihua and Li, Yuhan and Luo, Guan and Wang, Shuguang and Wu, Qianyi and Yang, Fan and Zhang, Junyang and Zhang, Xuanmeng},
  journal       = {arXiv preprint arXiv:2510.19944},
  year          = {2025},
  eprint        = {2510.19944},
  archivePrefix = {arXiv},
  primaryClass  = {cs.CV},
  doi           = {10.48550/arXiv.2510.19944},
  url           = {https://arxiv.org/abs/2510.19944}
}

@INPROCEEDINGS{cai2025computervisionfoundationmodels,
  author={Cai, Yancheng and Yin, Fei and Hammou, Dounia and Mantiuk, Rafal},
  booktitle={2025 IEEE/CVF Conference on Computer Vision and Pattern Recognition (CVPR)}, 
  title={Do computer vision foundation models learn the low-level characteristics of the human visual system?}, 
  year={2025},
  volume={},
  number={},
  pages={20039-20048},
  doi={10.1109/CVPR52734.2025.01866}}

\clearpage

\appendix

\section{Statements}
\label{appendix:statements}

\minisection{Limitations.}
While \ourmethod mitigates detail attenuation and enhances fine-grained texture fidelity for image-to-3D generation, several limitations remain. 
The framework is validated primarily on flow-matching-based native 3D diffusion pipelines (e.g., the TRELLIS family), and its applicability to other 3D generation paradigms remains to be evaluated. 
Moreover, \ourmethod primarily targets texture enhancement, and its current implementation uses DINO-series encoders; adaptation to other visual encoders may require re-tuning.

\minisection{Broader Impacts.}
Our proposed \ourmethod delivers substantial positive broader impacts by boosting the quality and practicality of image-to-3D generation. 
It lowers the technical barrier and production cost for creating high-fidelity textured 3D assets, supporting diverse applications such as game development, film and television virtual production, industrial design, cultural heritage digitization, robotics, and metaverse content creation, benefiting both professional creators and non-expert users. 
On the other hand, the strong detail-preserving capability may enable unauthorized duplication, counterfeiting, or misuse of copyrighted objects, branded products, or real-world artifacts, raising intellectual property and ethical concerns. 
To address such risks, responsible deployment, content authentication mechanisms, and clear usage norms are needed to ensure the technology benefits society while reducing potential harms.

\minisection{Ethical Statement.}
We acknowledge the potential ethical implications associated with generative image-to-3D technologies, including risks related to privacy, impersonation, and misuse of synthetic 3D assets. 
All models used in this work are trained on publicly available datasets and follow the usage policies of those datasets.
To promote transparency and responsible research, we will release the implementation details necessary to reproduce our results.
We encourage researchers and practitioners to use \ourmethod responsibly and to consider the broader societal implications when deploying image-to-3D synthesis.

\minisection{Reproducibility Statement.}
To ensure reproducibility, we will release the source code and inference and evaluation scripts required to reproduce the experimental results reported in this paper after the peer review process. 
All experiments are conducted using publicly available datasets, and detailed descriptions of the model architecture, inference configuration, and evaluation procedures are provided in the main paper and appendices.

\section{Implementation and Evaluation Details}
\label{appendix:implementation}

To validate the applicability of our method \ourmethod to existing image-to-3D models, we adopt TRELLIS \cite{xiang2025structured_trellis}, TRELLIS.2 \cite{xiang2025trellis2} and Hunyuan3D-2.1~\cite{hunyuan3d2025hunyuan3d_v21} as our backbone models, which allows us to evaluate the performance of \ourmethod when applied to different configurations of the DINO model and diverse 3D latent representations. 
Note that TRELLIS and Hunyuan3D-2.1 employ DINOv2~\cite{oquab2023dinov2} as its image encoder, while TRELLIS.2 adopts DINOv3~\cite{simeoni2025dinov3} for image encoding. 
These encoders are characterized by distinct 3D latent representations for 3D asset modeling. 
Evaluations conducted on these three backbones to demonstrate the generalizability of \ourmethod across diverse scenarios.
Below, we include the implementation and evaluation details of our methods and comparison methods.

\subsection{Evaluation Datasets}
\label{appendix:eval_dataset}

We evaluate \ourmethod on two public evaluation datasets: 3D-Arena~\cite{3d-arena} and Toys4K~\cite{stojanov2021using_toys4k}. 
3D-Arena is a public benchmark for image-to-3D generation, containing diverse object-centric reference images and providing a standardized testbed for comparing representative image-to-3D methods under a unified evaluation protocol. 
We evaluate all compared methods on the complete 3D-Arena evaluation set without sample-level filtering or post-hoc exclusion. All aggregate results are recomputed over the same complete set.

Toys4K is a category-diverse 3D object dataset containing 4,179 objects across 105 categories. 
We construct a category-balanced evaluation subset by randomly sampling two distinct objects without replacement from each category, yielding 210 objects in total. 
The sampling is performed once using a fixed random seed of 2026, and the resulting sample manifest is fixed before generation and shared across all compared methods.

By conducting evaluations on both datasets, we assess \ourmethod performance under two distinct settings: two general public benchmarks, 3D-Arena benchmark and Toys4K benchmark. 
For fair comparison, all quantitative evaluations are conducted on the common subset for which precomputed results from all compared methods are available.

\subsection{Evaluation Metrics}
\label{appendix:eval_metrics}

We adopt a comprehensive set of metrics to evaluate the generated 3D assets from complementary perspectives.
This section describes all metrics reported in the main table and the full appendix tables.
The main comparison focuses on CLIP-I~\cite{radford2021clip}, LPIPS~\cite{zhang2018lpips}, PSNR, SSIM~\cite{wang2004ssim}, ULIP-2~\cite{xue2024ulip}, and Uni3D~\cite{zhou2023uni3d}, covering rendered-view alignment, perceptual similarity, pixel-level fidelity, local structural similarity, and 3D-aware image-to-asset alignment.
Here, we additionally report CLIP-N, DINO, CLIP-FID, Edge Density, and Laplacian Variance.

Let \(I\) denote the input image, \(R_v\) and \(N_v\) the RGB and normal-map renderings from view \(v\), and \(V=12\) the number of rendered views.

\textit{3D-aware alignment metrics.}
These metrics evaluate generated assets from a 3D representation perspective by converting each mesh into a 10,000-point colored point cloud.
\begin{itemize}[leftmargin=*]
    \item \textit{ULIP-2~\cite{xue2024ulip} and Uni3D~\cite{zhou2023uni3d}.}
    ULIP-2 and Uni3D are models designed to understand and align 3D content with text/image.
    We first convert each generated mesh into a 10,000-point colored point cloud.
    Specifically, we sample surface points from the mesh and apply Farthest Point Sampling to obtain xyzrgb points.
    Before sampling, the point coordinates are normalized as
    \[
        \widetilde{\mathbf{p}}_i
        =
        \frac{\mathbf{p}_i-\bar{\mathbf{p}}}
        {\max_j\|\mathbf{p}_j-\bar{\mathbf{p}}\|_2},
        \qquad
        \bar{\mathbf{p}}
        =
        \frac{1}{M}\sum_{i=1}^{M}\mathbf{p}_i.
    \]
    The RGB value of each point is assigned by querying the available texture map, with vertex colors or face colors used as fallbacks when texture maps are unavailable.
    This colored point cloud is then fed into the ULIP-2 and Uni3D models to compute a similarity score against the image prompt:
    \[
        s_{\mathrm{3D}}(I,P)
        =
        \frac{
            f_{\mathrm{img}}(I)^\top f_{\mathrm{pc}}(P)
        }{
            \|f_{\mathrm{img}}(I)\|_2
            \|f_{\mathrm{pc}}(P)\|_2
        }.
    \]
    These scores provide a quantitative measure of how well the generated asset aligns with the condition from a native 3D perspective~\cite{xiang2025trellis2}.
\end{itemize}

\textit{Rendered-view texture metrics.}
These metrics evaluate visible texture by rendering each generated asset from 12 predefined viewpoints.
Specifically, we use four yaw angles of \(0^\circ\), \(90^\circ\), \(180^\circ\), and \(270^\circ\), combined with three pitch angles of \(25^\circ\), \(50^\circ\), and \(80^\circ\), and compare the rendered RGB views with the input image.

\begin{itemize}[leftmargin=*]
    \item \textit{CLIP-I~\cite{radford2021clip}.}
    We compute the CLIP cosine similarity between the input image and each rendered RGB view:
    \[
        s_v^{\mathrm{CLIP\text{-}I}}
        =
        \frac{
            f_{\mathrm{CLIP}}(I)^\top f_{\mathrm{CLIP}}(R_v)
        }{
            \|f_{\mathrm{CLIP}}(I)\|_2
            \|f_{\mathrm{CLIP}}(R_v)\|_2
        }.
    \]
    CLIP-I (Best view) and CLIP-I are respectively
    \[
        s_{\mathrm{best}}^{\mathrm{CLIP\text{-}I}}
        =
        \max_{1\leq v\leq V}s_v^{\mathrm{CLIP\text{-}I}},
        \qquad
        s_{\mathrm{all}}^{\mathrm{CLIP\text{-}I}}
        =
        \frac{1}{V}\sum_{v=1}^{V}s_v^{\mathrm{CLIP\text{-}I}}.
    \]

    \item \textit{DINO~\cite{caron2021dino}.}
    We compute the cosine similarity between DINO features of the input image and rendered RGB views:
    \[
        s_v^{\mathrm{DINO}}
        =
        \frac{
            f_{\mathrm{DINO}}(I)^\top f_{\mathrm{DINO}}(R_v)
        }{
            \|f_{\mathrm{DINO}}(I)\|_2
            \|f_{\mathrm{DINO}}(R_v)\|_2
        }.
    \]
    The best-view and all-view scores are
    \[
        s_{\mathrm{best}}^{\mathrm{DINO}}
        =
        \max_{1\leq v\leq V}s_v^{\mathrm{DINO}},
        \qquad
        s_{\mathrm{all}}^{\mathrm{DINO}}
        =
        \frac{1}{V}\sum_{v=1}^{V}s_v^{\mathrm{DINO}}.
    \]

    \item \textit{LPIPS~\cite{zhang2018lpips}.}
    We compute the perceptual distance between the input image and each rendered RGB view:
    \[
        d_v^{\mathrm{LPIPS}}
        =
        \sum_l
        \frac{1}{H_lW_l}
        \sum_{h,w}
        \left\|
            \mathbf{w}_l\odot
            \left(
                \widehat{\phi}_l(I)_{hw}
                -
                \widehat{\phi}_l(R_v)_{hw}
            \right)
        \right\|_2^2,
    \]
    where \(\widehat{\phi}_l\) denotes the channel-normalized feature at layer \(l\).
    LPIPS (Best view) and LPIPS are
    \[
        d_{\mathrm{best}}^{\mathrm{LPIPS}}
        =
        \min_{1\leq v\leq V}d_v^{\mathrm{LPIPS}},
        \qquad
        d_{\mathrm{all}}^{\mathrm{LPIPS}}
        =
        \frac{1}{V}\sum_{v=1}^{V}d_v^{\mathrm{LPIPS}}.
    \]
    Since LPIPS is sensitive to viewpoint, background, scale, and crop alignment, it is used as an auxiliary perceptual diagnostic.

    \item \textit{PSNR.}
    Peak signal-to-noise ratio is computed from the full-image RGB mean squared error:
    \[
        \operatorname{MSE}_v
        =
        \frac{1}{3HW}
        \sum_{c=1}^{3}\sum_{h=1}^{H}\sum_{w=1}^{W}
        \left(I_{h,w,c}-R_{v,h,w,c}\right)^2,
    \]
    \[
        \operatorname{PSNR}_v
        =
        10\log_{10}
        \left(
            \frac{1}{\operatorname{MSE}_v}
        \right).
    \]
    Its best-view and all-view variants are
    \[
        \operatorname{PSNR}_{\mathrm{best}}
        =
        \max_{1\leq v\leq V}\operatorname{PSNR}_v,
        \qquad
        \operatorname{PSNR}_{\mathrm{all}}
        =
        \frac{1}{V}\sum_{v=1}^{V}\operatorname{PSNR}_v.
    \]

    \item \textit{SSIM~\cite{wang2004ssim}.}
    Structural similarity is computed as
    \[
        \operatorname{SSIM}_v
        =
        \frac{
            (2\mu_I\mu_{R_v}+C_1)
            (2\sigma_{I,R_v}+C_2)
        }{
            (\mu_I^2+\mu_{R_v}^2+C_1)
            (\sigma_I^2+\sigma_{R_v}^2+C_2)
        },
    \]
    where
    \[
        C_1=(0.01)^2,
        \qquad
        C_2=(0.03)^2.
    \]
    The best-view and all-view variants are
    \[
        \operatorname{SSIM}_{\mathrm{best}}
        =
        \max_{1\leq v\leq V}\operatorname{SSIM}_v,
        \qquad
        \operatorname{SSIM}_{\mathrm{all}}
        =
        \frac{1}{V}\sum_{v=1}^{V}\operatorname{SSIM}_v.
    \]
    Like LPIPS, both PSNR and SSIM are sensitive to viewpoint, background, scale, and crop alignment, and are therefore used as auxiliary image-space diagnostics rather than standalone measures of 3D generation quality.

    \item \textit{CLIP-FID~\cite{jayasumana2024rethinking_fid,radford2021clip}.}
    We compute a distribution-level Fréchet distance in CLIP feature space between input images and CLIP-I-selected best-view renderings:
    \[
        \operatorname{CLIP\text{-}FID}
        =
        \|\boldsymbol{\mu}_I-\boldsymbol{\mu}_R\|_2^2
        +
        \operatorname{Tr}
        \left(
            \boldsymbol{\Sigma}_I
            +
            \boldsymbol{\Sigma}_R
            -
            2
            (\boldsymbol{\Sigma}_I\boldsymbol{\Sigma}_R)^{1/2}
        \right),
    \]
    where \(\boldsymbol{\mu}_I,\boldsymbol{\Sigma}_I\) and
    \(\boldsymbol{\mu}_R,\boldsymbol{\Sigma}_R\) denote the means and covariance matrices of the corresponding CLIP features.
\end{itemize}

\textit{Normal-map and geometric metrics.}
These metrics evaluate normal consistency and rendered geometric responses.

\begin{itemize}[leftmargin=*]
    \item \textit{CLIP-N.}
    We render normal maps from the same 12 viewpoints and compute
    \[
        s_v^{\mathrm{CLIP\text{-}N}}
        =
        \frac{
            f_{\mathrm{CLIP}}(I)^\top f_{\mathrm{CLIP}}(N_v)
        }{
            \|f_{\mathrm{CLIP}}(I)\|_2
            \|f_{\mathrm{CLIP}}(N_v)\|_2
        }.
    \]
    CLIP-N (Best view) and CLIP-N are
    \[
        s_{\mathrm{best}}^{\mathrm{CLIP\text{-}N}}
        =
        \max_{1\leq v\leq V}s_v^{\mathrm{CLIP\text{-}N}},
        \qquad
        s_{\mathrm{all}}^{\mathrm{CLIP\text{-}N}}
        =
        \frac{1}{V}\sum_{v=1}^{V}s_v^{\mathrm{CLIP\text{-}N}}.
    \]

    \item \textit{Edge Density~\cite{rosenholtz2007visual}.}
    Let \(E_v(p)\in\{0,1\}\) denote whether pixel \(p\) is detected as an edge in rendered view \(v\). Edge Density is
    \[
        \operatorname{ED}_v
        =
        \frac{1}{HW}
        \sum_{p=1}^{HW}E_v(p).
    \]
    Its all-view value is
    \[
        \operatorname{ED}_{\mathrm{all}}
        =
        \frac{1}{V}\sum_{v=1}^{V}\operatorname{ED}_v,
    \]
    while \(\operatorname{ED}_{\mathrm{best}}\) is evaluated on the CLIP-I-selected best-view rendering.

    \item \textit{Laplacian Variance~\cite{pech2000diatom,pertuz2013analysis}.}
    Let \(G_v\) be the grayscale rendering and \(\Delta G_v\) its Laplacian response. Laplacian Variance is
    \[
        \operatorname{LV}_v
        =
        \frac{1}{HW}
        \sum_{p=1}^{HW}
        \left(
            \Delta G_v(p)
            -
            \overline{\Delta G_v}
        \right)^2,
    \]
    where
    \[
        \overline{\Delta G_v}
        =
        \frac{1}{HW}
        \sum_{p=1}^{HW}\Delta G_v(p).
    \]
    Its all-view value is
    \[
        \operatorname{LV}_{\mathrm{all}}
        =
        \frac{1}{V}\sum_{v=1}^{V}\operatorname{LV}_v,
    \]
    while \(\operatorname{LV}_{\mathrm{best}}\) is evaluated on the CLIP-I-selected best-view rendering.
\end{itemize}

\section{Additional Experimental Results}

\subsection{High-frequency Perception}
\begin{figure}[t]
\centering
\includegraphics[width=0.75\linewidth]{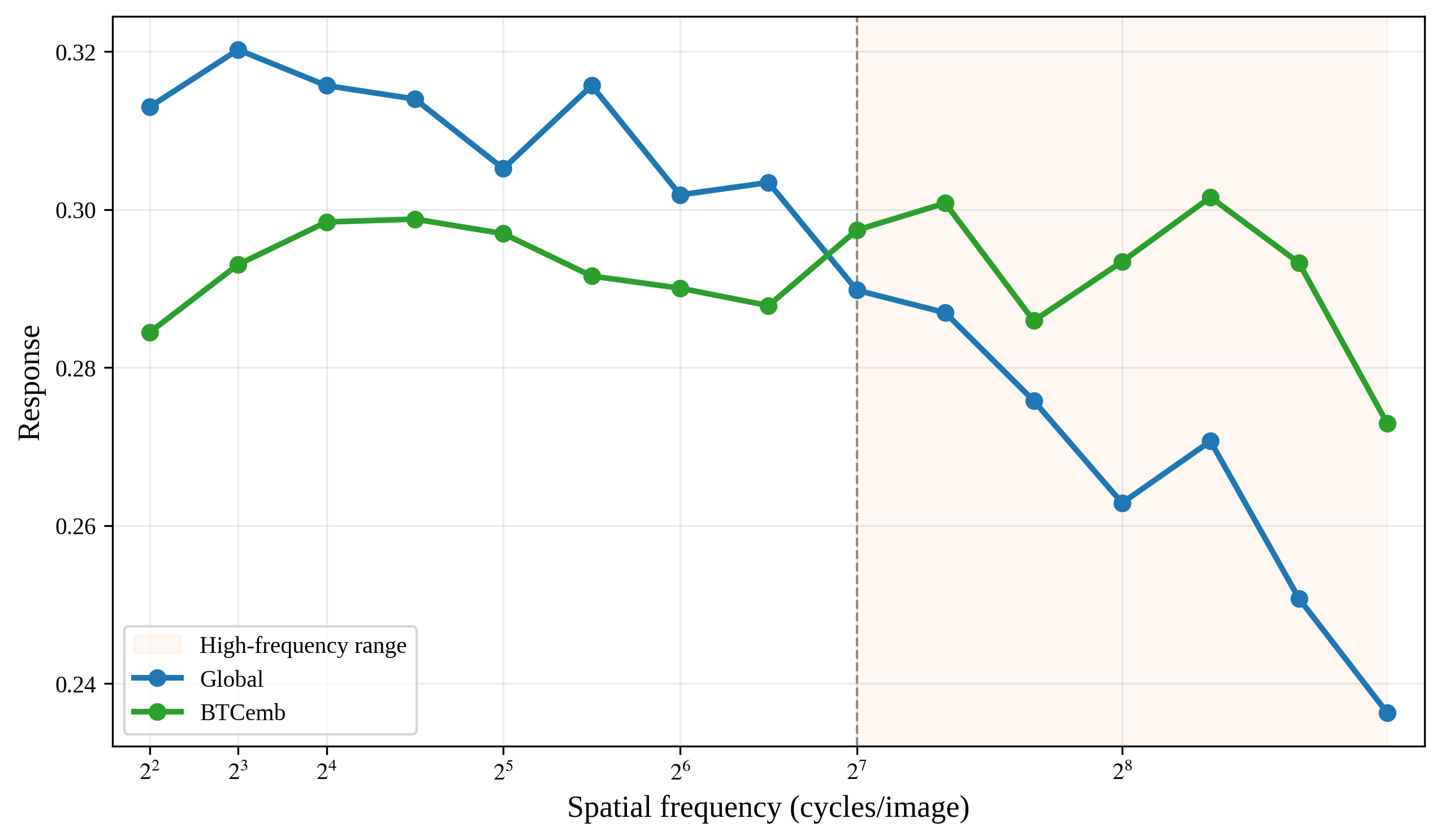}
\caption{\textbf{Frequency response of Global and \btcembed.} We compare full-field sinusoidal gratings with matched flat images using the normalized angular distance between their flattened complete conditioning tensors. Each point averages eight orientation--phase configurations. The shaded region highlights $128$--$448$ cycles/image, and sampled frequencies are displayed at equal spacing.}
\label{fig:high_frequency_response}
\end{figure}

Detailed textures and sharp boundaries often involve high-frequency image components, making sensitivity to such components relevant to local detail preservation. To complement the generation results, we compare the frequency responses of Global and \btcembed to examine how local tile encoding affects the representation of high-frequency information.

To examine sensitivity to high-frequency details at the encoding stage, we probe the frozen DINOv3 encoder~\cite{simeoni2025dinov3} using controlled $1024\times1024$ full-field sinusoidal gratings. Stimuli are generated in linear luminance with a background level of $0.25$ and Michelson contrast of $0.25$, using a raised-cosine taper over the outer $8\%$ of each image dimension, and then converted to sRGB while retaining float32 precision. The displayed frequencies are $4,8,16,24,32,48,64,96,128,160,192,256,320,384,$ and $448$ cycles/image. At each frequency, we evaluate four orientations $\theta\in\{0^\circ,45^\circ,90^\circ,135^\circ\}$ and two phases $\phi\in\{0,\pi/2\}$.

Global uses the complete conditioning tensor obtained by encoding the full image. For \btcembed, tiles are resized to the encoder input resolution, encoded independently, and combined through weighted aggregation of their complete conditioning tensors. The tile extraction and aggregation settings remain fixed throughout the experiment.

Let $I_{f,\theta,\phi}$ denote a grating and $I_0$ its uniform reference at the same background luminance. For each method $m$, let $z_m(I)$ denote the flattened complete conditioning tensor, retaining all tokens. Both the stimulus and its reference undergo the same method-specific encoding and aggregation. We measure the response using the normalized angular distance~\cite{cai2025computervisionfoundationmodels}, averaged over the eight orientation--phase configurations:
\begin{equation} r_m(f)=\frac{1}{8}\sum_{\theta,\phi}\frac{1}{\pi}\arccos\!\left(\frac{z_m(I_{f,\theta,\phi})^\top z_m(I_0)}{\|z_m(I_{f,\theta,\phi})\|_2\,\|z_m(I_0)\|_2}\right). \label{eq:high_frequency_response} \end{equation}
A larger response indicates greater separation between the grating and its flat reference in the complete conditioning representation. Responses are computed for each orientation--phase configuration before averaging.

As shown in~\Cref{fig:high_frequency_response}, Global responds more strongly at the lower sampled frequencies, whereas \btcembed maintains stronger responses toward the high-frequency end, particularly at $256$--$448$ cycles/image. This suggests that tiled encoding and aggregation help preserve sensitivity to fine spatial variations that are less distinguishable under global encoding. Since each tile is resized to the encoder input resolution, this behavior reflects the combined effects of the local field of view, rescaling, and feature aggregation. These results provide encoder-level evidence complementary to the region-level and rendered-image evaluations, rather than a direct measurement of generated texture fidelity.

\subsection{Image Feature Additivity and Same-Category Compatibility}
\label{app:feature_additivity_same_category}

To further analyze the composability and compatibility of local features, we evaluate global-local features cosine distance on the fixed 210-object Toys4K set spanning all 105 categories.

We additionally use same-category hard negatives to determine whether the result is driven only by coarse category separation. For this analysis, we use 210 objects, with two objects from each of the 105 Toys4K categories. Category folders are traversed in lexicographic order, and object sampling uses the fixed seed 2026.

Cosine distance is defined as
\begin{equation} s_{\mathrm{cos}}(a,b)=\cos(a,b). \end{equation}
The same-category cosine margin is defined as
\begin{equation} m_{\mathrm{same}}=s_{\mathrm{cos}}(f_i^l,f_i^g)-s_{\mathrm{cos}}(f_i^l,f_j^g), \end{equation}
where $f_i^l$ is the local representation, $f_i^g$ is its matched global representation, and $f_j^g$ is a non-matching global representation from the same category. A positive margin indicates that the local representation has higher cosine distance with its matched global representation than with another object from the same category. Higher similarity and margin therefore indicate stronger global-local compatibility and better preservation of input-specific semantic identity. Results are reported as mean [95\% bootstrap confidence interval].

\begin{table}[t]
\centering
\small
\setlength{\tabcolsep}{5pt}
\renewcommand{\arraystretch}{1.12}
\caption{Image feature additivity and same-category compatibility.}
\label{tab:feature_additivity_same_category}
\begin{tabular}{lcc}
\toprule
Representation
& \shortstack[c]{Global-local Features Cos. Dist. $\uparrow$}
& \shortstack[c]{Same-category cosine margin $\uparrow$} \\
\midrule
Average tiles
& 0.7466 [0.7179, 0.7758]
& 0.1728 [0.1433, 0.2041] \\
\textbf{BTCemb (Ours)}
& \textbf{0.8242 [0.8035, 0.8434]}
& \textbf{0.2195 [0.1894, 0.2517]} \\
\bottomrule
\end{tabular}
\end{table}

Both same-category margins remain positive. \textbf{BTCemb} further increases the global-local features cosine distance and the same-category cosine margin compared with Average tiles. Since the matched and negative objects belong to the same category, these results show that the observed compatibility preserves input-specific semantic identity rather than reflecting only coarse category separation.

\begin{figure}[t]
\centering
\includegraphics[width=0.9\linewidth]{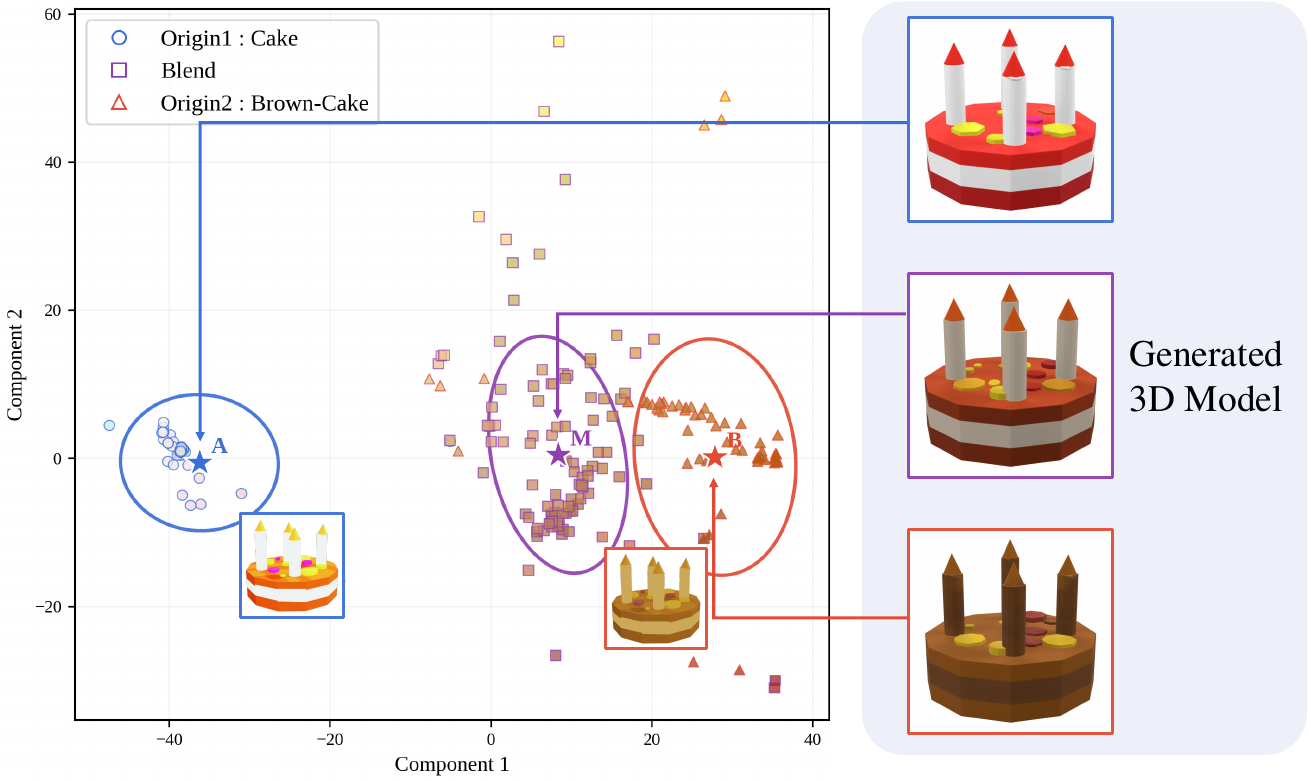}
\caption{\textbf{Qualitative analysis of image feature additivity.} Left: PCA visualization of sampled CIELAB surface colors from the generated models, with the two input images shown as insets. Right, from top to bottom: results conditioned on the original, blended, and color-modified image embeddings.}
\label{fig:appendix_additivity}
\end{figure}

To complement the feature-space analysis above, we further examine whether blended image embeddings can be directly used for image-to-3D generation. We independently encode an original rendered image and its color-modified version, and use the equal-weight average of their complete conditioning embeddings to condition the frozen TRELLIS.2 generator. As shown in~\Cref{fig:appendix_additivity}, the generated result combines color characteristics from both inputs. This example provides additional qualitative support for \textit{image feature additivity} and the compatibility of blended embeddings with the pretrained generator.

\subsection{Runtime Overhead}
\label{app:runtime}

\begin{table}[t]
\centering
\small
\setlength{\tabcolsep}{5pt}
\caption{Runtime comparison of baseline pipelines before and after integrating our method on 3D-Arena~\cite{3d-arena}.
All runtimes are measured on a single NVIDIA RTX PRO 6000 GPU.
}
\label{tab:runtime_comparison}
\begin{tabular}{lcccc}
\toprule
Method &Sampling Steps & Baseline Runtime (s) & Ours Runtime (s) & Overhead (s) \\
\midrule
Hunyuan3D-2.1 & 50 & $76.27 \pm 1.37$ & $78.63 \pm 1.13$ & $2.36 \pm 1.59$ \\
TRELLIS       & 20 & $2.96 \pm 0.23$  & $4.89 \pm 0.24$  & $1.93 \pm 0.11$ \\
TRELLIS.2     & 12 & $15.88 \pm 1.43$ & $18.30 \pm 1.44$ & $2.41 \pm 1.25$ \\
\bottomrule
\end{tabular}
\end{table}

Table~\ref{tab:runtime_comparison} reports the runtime of each baseline pipeline and its variant integrated with our method on 3D-arena~\cite{3d-arena}, presented as mean $\pm$ standard deviation over three runs.
The reported time includes preprocessing, feature extraction, and the full pipeline execution, while excluding asset export due to its relatively large variance across runs, and runtime is measured with \texttt{time.perf\_counter()}.
Note that the reported runtime is mainly intended to compare each pipeline before and after integrating our method, since the default sampling schedules and implementations differ across methods.
Each pipeline is evaluated with its default sampling schedule, i.e., 50 sampling steps for Hunyuan3D-2.1, 20 sampling steps for TRELLIS, and 12 sampling steps for TRELLIS.2.

After adopting \ourmethod, the overhead mainly comes from feature extraction.
Specifically, the image encoders used for feature construction are DINOv2-giant for Hunyuan3D-2.1, DINOv2-large for TRELLIS, and DINOv3-ViT-L/16 for TRELLIS.2.
Overall, the computational overhead introduced by adopting our method is negligible, especially for computationally intensive generation pipelines.

\subsection{Additional Quantitative Results}
\label{appendix:additional_quantitative_results}

Table~\ref{tab:appendix_all_additional_metrics} reports the supplementary evaluation results on 3D-Arena and Toys4K.
These metrics complement the main comparison by measuring normal-map alignment with CLIP-N, distribution-level fidelity with CLIP-FID, rendered-view feature consistency with DINO, perceptual similarity with LPIPS (Best view), and rendered detail responses with Edge Density and Laplacian Variance.

Overall, \ourmethod shows consistent benefits on several supplementary diagnostics across different backbones.
On 3D-Arena, applying \ourmethod to TRELLIS and TRELLIS.2 improves distribution-level fidelity and rendered-view feature consistency, as reflected by lower CLIP-FID and higher DINO scores.
It also improves several perceptual and detail-related indicators, including LPIPS (Best view) and Laplacian Variance for TRELLIS-family backbones.
For Hunyuan3D-2.1, \ourmethod also improves CLIP-FID and DINO, suggesting that the proposed refinement is not limited to TRELLIS-family models but can also serve as a plug-and-play method for other strong image-to-3D backbones.

On Toys4K dataset, the advantages of \ourmethod are more evident on texture-sensitive diagnostics.
For Hunyuan3D-2.1, \ourmethod improves CLIP-FID, DINO, LPIPS (Best view), and Edge Density, indicating better rendered appearance fidelity and richer local detail responses.
For TRELLIS and TRELLIS.2, \ourmethod also improves multiple supplementary metrics, especially DINO, LPIPS (Best view), Edge Density, and Laplacian Variance.

No-reference detail metrics should be interpreted with caution.
The high Laplacian Variance of 3DTopia-XL on Toys4K is mainly driven by sharp silhouettes, hard geometric boundaries, and artifact-like edges, rather than faithful texture recovery.
Thus, Edge Density and Laplacian Variance are used only as auxiliary rendered-detail diagnostics.

\begin{table*}[t]
\centering
\tiny
\setlength{\tabcolsep}{1.8pt}
\renewcommand{\arraystretch}{1.12}
\caption{\textbf{Additional semantic, distribution, perceptual, and detail metrics on 3D-Arena and Toys4K.} Results are averaged over the corresponding evaluation sets. BV denotes best-view results, and E-Den denotes the Edge Density metric. Higher is better for CLIP-N, DINO, Edge Density, and Laplacian Variance, while lower is better for CLIP-FID and LPIPS. Best results are shown in \textbf{bold} and second-best results are \underline{underlined}. For shape-only or untextured methods, DINO scores are not reported for shape-only or untextured methods.}
\label{tab:appendix_all_additional_metrics}
\resizebox{\textwidth}{!}{
\begin{tabular}{l c ccccc ccccc}
\toprule[1pt]
Method
& \shortstack[c]{Train\\Free}
& \shortstack[c]{CLIP-N\\BV $\uparrow$}
& \metric{CLIP-N}{\uparrow}
& \metric{CLIP-FID}{\downarrow}
& \shortstack[c]{DINO\\BV $\uparrow$}
& \metric{DINO}{\uparrow}
& \shortstack[c]{LPIPS\\BV $\downarrow$}
& \shortstack[c]{E-Den \\BV $\uparrow$}
& \metric{E-Den}{\uparrow}
& \shortstack[c]{Lap. Var.\\BV $\uparrow$}
& \metric{Lap. Var.}{\uparrow} \\
\midrule

\multicolumn{12}{c}{3D-Arena Benchmark} \\
\midrule
TripoSG & \xmark
& \textbf{0.5725} & \textbf{0.4464} & 40.4435 & -- & --
& -- & 0.0106 & 0.0099 & 98.9921 & 93.9380 \\

Hi3DGen & \xmark
& \underline{0.5693} & \underline{0.4455} & 39.9314 & -- & --
& -- & 0.0120 & 0.0108 & 99.4825 & 94.1028 \\

3DTopia-XL & \xmark
& 0.5030 & 0.3725 & 25.5023 & 0.6326 & 0.3095
& \textbf{0.3628} & 0.0171 & 0.0159 & 167.3116 & 157.6823 \\

\midrule

Hunyuan3D-2.1 & \xmark
& 0.5242 & 0.4034 & 29.3601 & 0.5890 & 0.3470
& 0.4102 & 0.0208 & 0.0181 & 149.1459 & 138.6304 \\

\rowcolor{gray!12}
\oursmethod & \cmark
& 0.5263 & 0.4044 & 28.7202 & 0.5940 & 0.3539
& 0.4083 & 0.0216 & 0.0188 & 152.0670 & 140.6304 \\

TRELLIS & \xmark
& 0.5507 & 0.4118 & 29.8881 & \underline{0.7000} & 0.3976
& 0.3858 & 0.0124 & 0.0117 & 178.2178 & \underline{172.8362} \\

\rowcolor{gray!12}
\oursmethod & \cmark
& 0.5510 & 0.4184 & \underline{25.2971} & \textbf{0.7137} & \textbf{0.4153}
& \underline{0.3796} & 0.0175 & 0.0163 & \underline{179.0849} & \textbf{173.4081} \\

TRELLIS.2 & \xmark
& 0.5158 & 0.4195 & 27.7110 & 0.6353 & 0.3842
& 0.3922 & \underline{0.0246} & \underline{0.0215} & 174.8840 & 148.0262 \\

\rowcolor{gray!12}
\oursmethod & \cmark
& 0.5465 & 0.4332 & \textbf{24.8679} & 0.6711 & \underline{0.4071}
& 0.3916 & \textbf{0.0276} & \textbf{0.0233} & \textbf{184.1118} & 165.6617 \\

\midrule[1pt]
\multicolumn{12}{c}{Toys4K Benchmark} \\
\midrule

TripoSG & \xmark
& \textbf{0.6205} & \textbf{0.4854} & 39.3412 & -- & --
& -- & 0.0140 & 0.0127 & 123.2876 & 113.0368 \\

Hi3DGen & \xmark
& 0.6170 & \underline{0.4839} & 39.2177 & -- & --
& -- & 0.0135 & 0.0127 & 115.5724 & 110.7903 \\

3DTopia-XL & \xmark
& 0.5396 & 0.4103 & 31.0932 & 0.6057 & 0.3187
& 0.7266 & 0.0226 & 0.0202 & \textbf{352.5917} & \textbf{317.5117} \\

\midrule

Hunyuan3D-2.1 & \xmark
& 0.6081 & 0.4802 & \underline{23.7477} & \underline{0.7499} & \underline{0.4759}
& \underline{0.7136} & 0.0237 & \underline{0.0224} & 190.9429 & 179.7303 \\

\rowcolor{gray!12}
\oursmethod & \cmark
& \underline{0.6176} & 0.4804 & \textbf{23.5111} & \textbf{0.7539} & \textbf{0.4786}
& \textbf{0.7129} & \textbf{0.0248} & \textbf{0.0231} & 191.0055 & 180.0754 \\

TRELLIS & \xmark
& 0.5961 & 0.4626 & 25.9937 & 0.7184 & 0.4520
& 0.7221 & 0.0158 & 0.0149 & 172.4711 & 168.4326 \\

\rowcolor{gray!12}
\oursmethod & \cmark
& 0.5988 & 0.4651 & 25.4794 & 0.7291 & 0.4583
& 0.7213 & 0.0164 & 0.0154 & 179.9012 & 172.8051 \\

TRELLIS.2 & \xmark
& 0.6093 & 0.4819 & 26.1943 & 0.7021 & 0.4612
& 0.7278 & 0.0237 & 0.0217 & 210.2709 & 193.2877 \\

\rowcolor{gray!12}
\oursmethod & \cmark
& 0.6135 & 0.4832 & 26.0801 & 0.7072 & 0.4625
& 0.7274 & \underline{0.0243} & 0.0222 & \underline{225.4874} & \underline{206.3683} \\

\bottomrule[1pt]
\end{tabular}
}
\end{table*}

To complement global alignment metrics, we conduct a region-level paired evaluation to examine local appearance fidelity, including the preservation of fine textures and boundaries. Table~\ref{tab:region_level_paired_metrics} reports results on Toys4K using a fixed $4\times4$ grid defined solely by the GT alpha mask. The evaluation covers 210 objects, 12 matched views per object, and 4,903 valid regions at $1024\times1024$ resolution. Region PSNR and SSIM assess pixel-level fidelity and structural similarity within individual regions, while paired win rates indicate how consistently these local measures improve across regions.

\begin{table*}[t]
\centering
\small
\setlength{\tabcolsep}{10pt}
\renewcommand{\arraystretch}{1.12}
\caption{\textbf{Region-level paired evaluation on Toys4K.} Region PSNR and SSIM are computed over spatially paired valid regions. Each merged win-rate cell reports the percentage of paired regions where \ourmethod outperforms the corresponding baseline. Win rates are estimated from 10,000 sampling trials and reported as estimate $\pm$ confidence-interval half-width. Higher is better.}
\label{tab:region_level_paired_metrics}

\begin{tabular}{l c c c c}
\toprule[1pt]
Method
& \shortstack[c]{Region PSNR $\uparrow$}
& \shortstack[c]{PSNR Win Rate}
& \shortstack[c]{Region SSIM $\uparrow$}
& \shortstack[c]{SSIM Win Rate} \\
\midrule

Hunyuan3D-2.1
& 19.6319
& 
& 0.8135
&  \\

\rowcolor{gray!12}
\oursmethod
& 20.4305
& 76.6\% $\pm$ 4.9\%
& 0.8309
& 73.5\% $\pm$ 5.7\% \\

\midrule

TRELLIS
& 19.2718
& 
& 0.8064
&  \\

\rowcolor{gray!12}
\oursmethod
& 19.8367
& 69.1\% $\pm$ 6.1\%
& 0.8205
& 71.1\% $\pm$ 5.4\% \\

\midrule

TRELLIS.2
& 19.8568
& 
& 0.8113
&  \\

\rowcolor{gray!12}
\oursmethod
& 20.4711
& 76.2\% $\pm$ 4.5\%
& 0.8470
& 75.1\% $\pm$ 3.8\% \\

\bottomrule[1pt]
\end{tabular}
\end{table*}

\begin{table*}[t]
\centering
\small
\setlength{\tabcolsep}{9pt}
\renewcommand{\arraystretch}{1.12}
\caption{\textbf{Region-level paired evaluation on Toys4K.} Edge Density (E-Den) and Laplacian Variance (Lap. Var.) provide auxiliary measures of edge richness and high-frequency responses in rendered images. Win rate reports the percentage of paired images where \ourmethod yields a higher metric value than the corresponding baseline. Win rates are computed from 10,000 sampling trials and reported as value $\pm$ the half-width of the 95\% confidence interval.}
\label{tab:image_level_paired_detail_metrics}

\begin{tabular}{l c c c c}
\toprule[1pt]
Method
& \shortstack[c]{Edge Density $\uparrow$}
& \shortstack[c]{E-Den Win Rate}
& \shortstack[c]{Lap. Var. $\uparrow$}
& \shortstack[c]{Lap. Var. Win Rate} \\
\midrule

Hunyuan3D-2.1
& 0.0214
&
& 188.9429
& \\

\rowcolor{gray!12}
\oursmethod
& 0.0278
& 73.1\% $\pm$ 4.7\%
& 195.0055
& 71.9\% $\pm$ 5.2\% \\

\midrule

TRELLIS
& 0.0159
&
& 175.3981
& \\

\rowcolor{gray!12}
\oursmethod
& 0.0194
& 67.7\% $\pm$ 5.6\%
& 180.9137
& 70.5\% $\pm$ 4.9\%\\

\midrule

TRELLIS.2
& 0.0234    
&
& 213.3255
& \\

\rowcolor{gray!12}
\oursmethod
& 0.0310
& 76.4\% $\pm$ 4.8\%
& 228.5734
& 77.1\% $\pm$ 5.1\% \\

\bottomrule[1pt]
\end{tabular}
\end{table*}

\subsection{Additional Qualitative Results}
\label{appendix:additional_qualitative_results}
\begin{figure}[t]
    \centering
    \includegraphics[width=\textwidth]
    {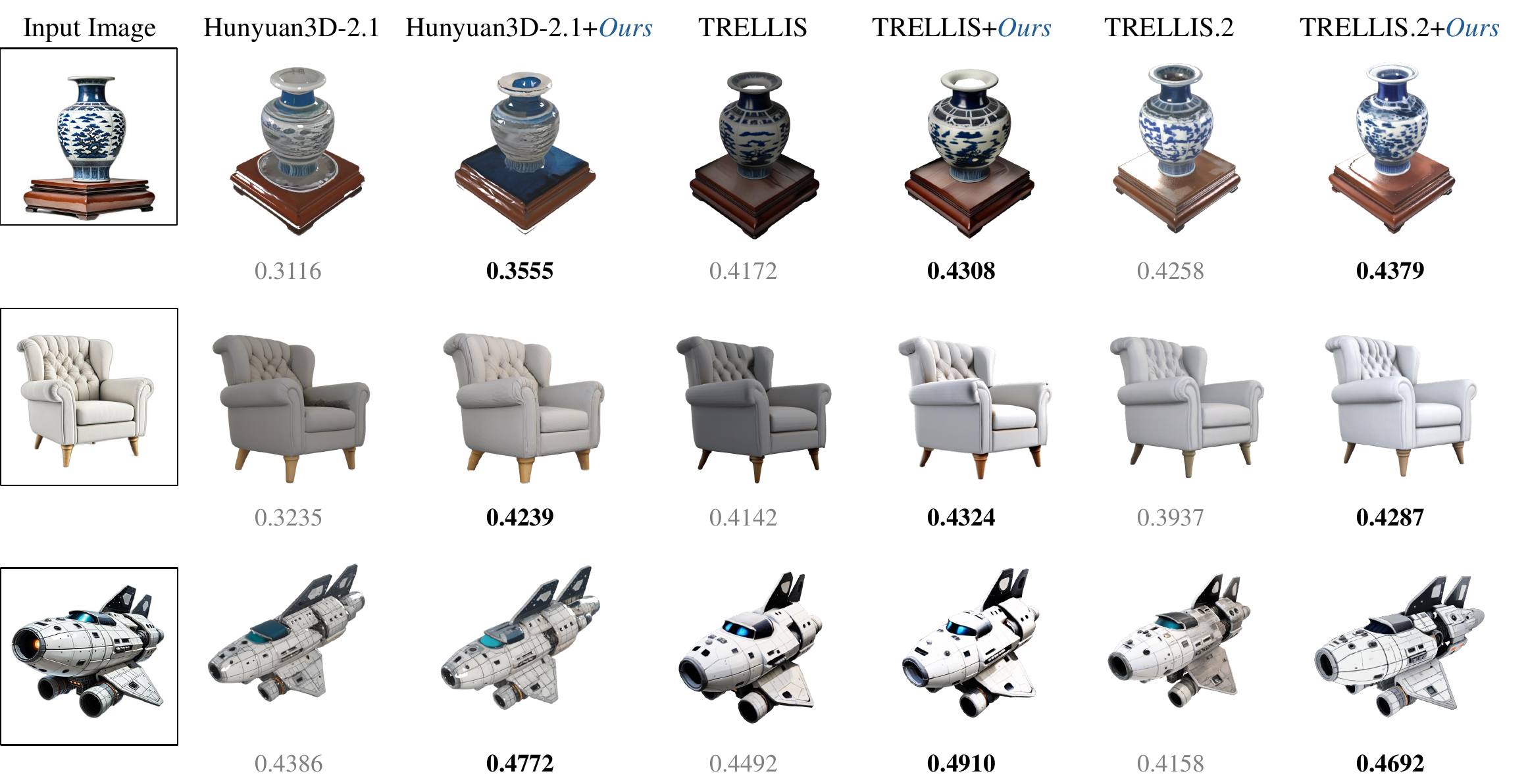}
    \caption{More qualitative comparisons by ULIP score with image-to-3D generation models.}
    \label{fig:quality_score}
\end{figure}

\begin{figure}[t]
\centering
\includegraphics[width=\textwidth]
{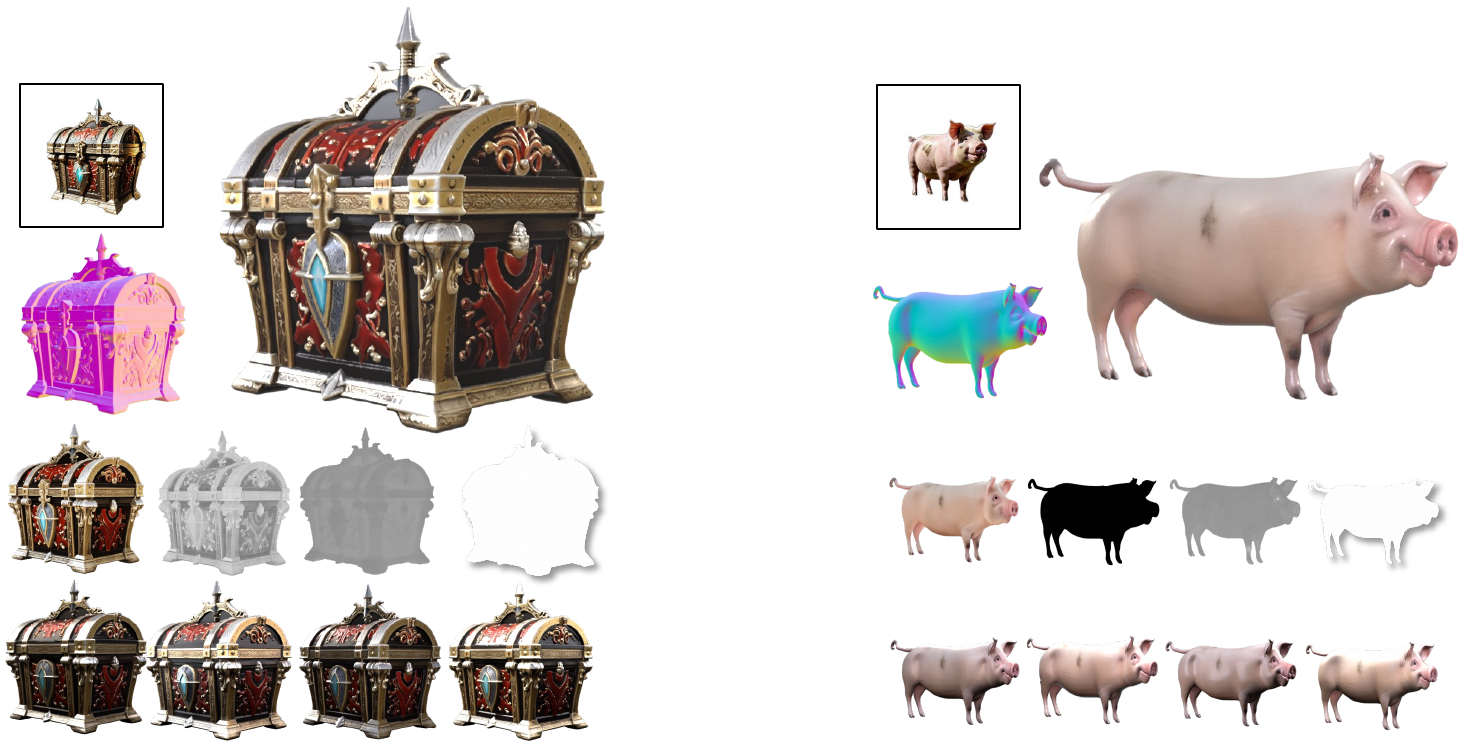}
\caption{Additional visualizations of \ourmethod.
For each example, the boxed image shows the input, the large image shows the final rendered result, and the colored image shows the corresponding normal map.
The eight smaller views below provide further material and appearance analysis: the first row shows the base color, metallic, roughness, and alpha maps, while the second row presents four renderings under different realistic illumination from the same viewpoint.}
    \label{fig:more_details}
\end{figure}

\begin{figure}[t]
\centering
\includegraphics[width=\textwidth]
{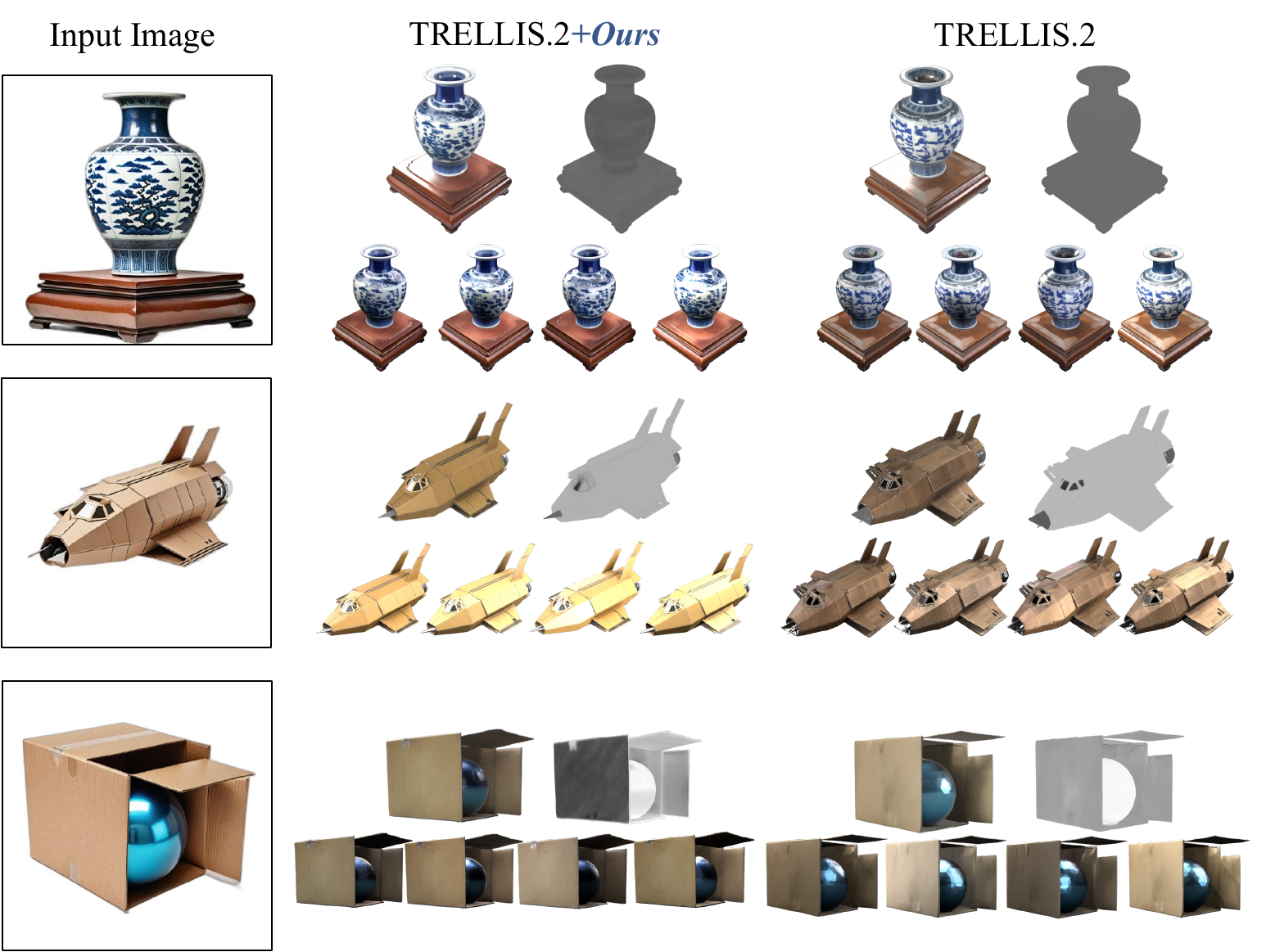}
\caption{Additional comparisons between TRELLIS.2+\ourmethod\ and TRELLIS.2.
For each input image, the left column shows the input, while the middle and right columns present the results of TRELLIS.2+\ourmethod\ and TRELLIS.2, respectively.
For each generated result, the left larger view shows the textured rendering, and the right larger view shows the corresponding material visualization.
The four smaller views below present renderings under different realistic lighting from the same viewpoint.}
    \label{fig:comp_to_t2}
\end{figure}

We provide additional qualitative results to further demonstrate the effectiveness of \ourmethod as a plug-in method for image-to-3D generation.
Figure~\ref{fig:quality_score} shows additional visualization obtained by integrating \ourmethod into different baseline models~\cite{hunyuan3d2025hunyuan3d_v21,xiang2025trellis2,xiang2025structured_trellis}. 
The ULIP scores~\cite{xue2024ulip} reported below each rendering show that \ourmethod consistently improves over the original baselines. 
These visual results demonstrate the generalization ability of our training-free plug-in method, which improves existing image-to-3D models by producing 3D assets with accurate geometry and high-fidelity textures without requiring additional training.

Direct visual comparisons between the original baseline method TRELLIS.2 \cite{xiang2025trellis2} and methods integrated with \ourmethod are presented in Figure~\ref{fig:comp_to_t2}. The integration of \ourmethod yields marked enhancement in the details of the final textures. This demonstrates the effectiveness of \ourmethod in preserving local features from the condition image while mitigating the multi-view inconsistency and over-smoothing artifacts prevalent in standard diffusion-based 3D generation.

To further illustrate the representational capacity of our approach, Figure~\ref{fig:more_details} shows high-fidelity 3D assets generated utilizing \ourmethod. The method accurately reconstructs intricate microscopic details, such as the complex carvings and metallic specularity of the treasure chest, as well as the subtle surface variations of the pig. The accompanying high-resolution normal maps and multi-view renderings validate the potential of \ourmethod in synthesizing high-quality 3D assets with complex material properties and geometrically consistent details.

\subsection{User Study}
\label{appendix:user_study}
As noted in~\ref{appendix:eval_metrics}, automatic evaluation metrics such as CLIP-I and ULIP-2 primarily measure global semantic alignment and may not fully capture fine-grained texture fidelity, local geometry coherence, or human-perceived visual quality. 
To complement the automated quantitative results in~\ref{appendix:additional_quantitative_results} and further validate \ourmethod for detail-preserving image-to-3D generation, we conduct a pairwise human preference study comparing 3D assets generated by the baseline TRELLIS.2 and its variant integrated with \ourmethod on 3D-Arena~\cite{3d-arena}.

\textbf{Stimuli Preparation.}
We use all 101 samples from the 3D-Arena dataset to construct the stimulus pool for the user study.
For each sample, we pair the GLB assets generated by the baseline TRELLIS.2 and by TRELLIS.2 integrated with \ourmethod.
Both assets are loaded directly into interactive 3D viewers and displayed side by side, with the conditioning input image positioned above them as a visual reference.
Within each pair, the assets are centered and displayed using a shared scale, identical lighting and background conditions, and matched initial camera settings.
Participants can freely rotate, zoom, and pan the models to inspect their geometry and appearance, with synchronized camera controls available to facilitate comparison.
Each participant evaluates 20 distinct pairs randomly sampled from the full 101-sample pool.

\textbf{Evaluation Protocol.}
The user study involved 20 participants and adopts a two-alternative forced-choice (2AFC) pairwise comparison protocol~\cite{Qiu2024richdreamer,liu2024one2345,zhang2018lpips}. Each participant evaluates 20 distinct samples randomly drawn without replacement from the full pool of 101 3D-Arena samples, with the presentation order randomized for each participant.
For each sample, the conditioning input image is displayed above two interactive 3D viewers labeled A and B, without revealing the corresponding generation methods. Participants can rotate, zoom, and pan the models, with synchronized camera controls available to facilitate comparison. The assignment of the baseline and \ourmethod outputs to A and B is randomized for each sample within each participant's session using a session-specific random seed, to mitigate positional bias.
Participants answer the question:\textit{``Which 3D asset is better?''} by selecting either \textit{``A is better''} or \textit{``B is better.''}
They are instructed to choose the asset that better matches the reference image, considering its shape, appearance, and visual details. Responses are recorded through a custom web-based interface. The interface uses a responsive layout to support both desktop and mobile devices, while retaining the same reference image, paired interactive 3D viewers, and response options.

\begin{figure}[t]
\centering
\includegraphics[width=\linewidth]{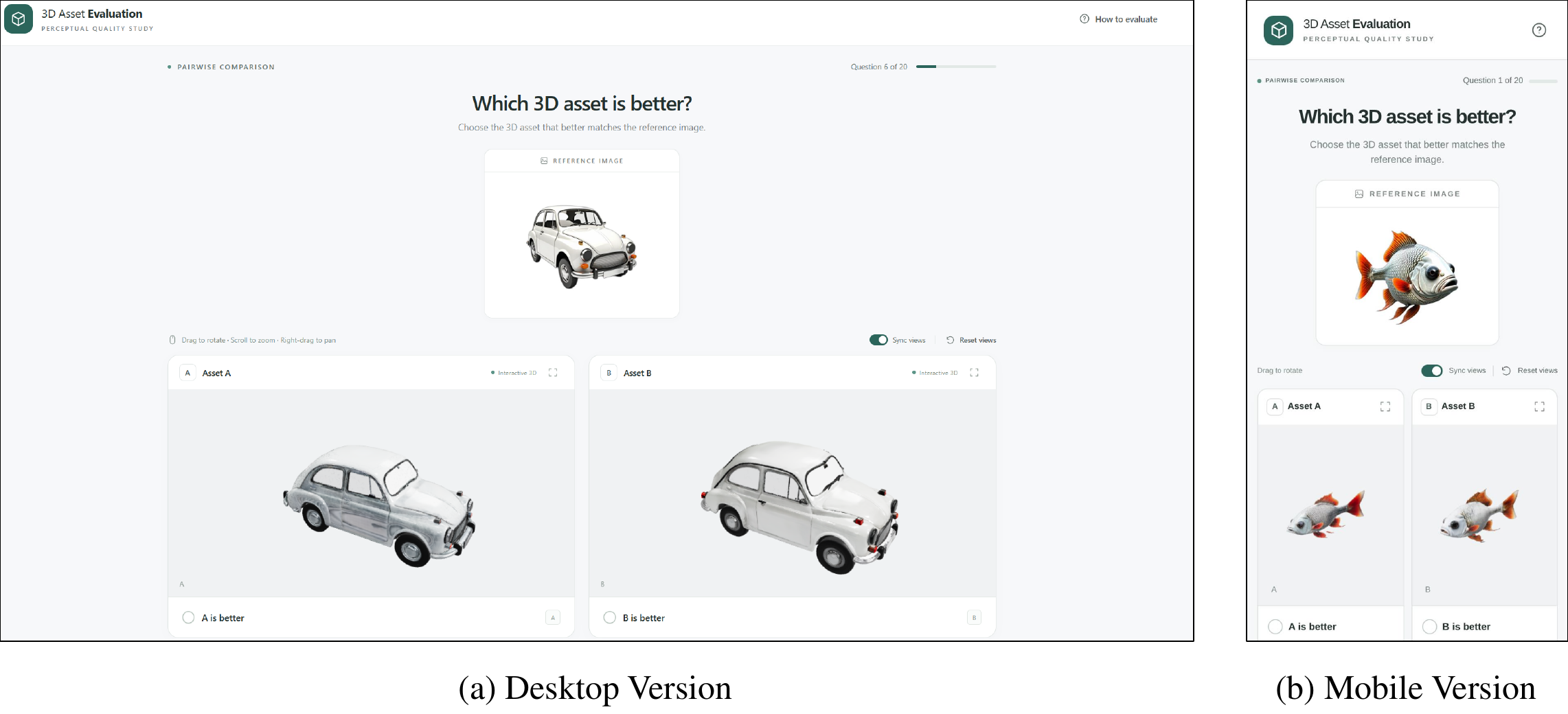}
\vspace{3mm}
\caption{
\textbf{User study interface.}
Desktop (left) and mobile (right) layouts of our web-based evaluation interface. The reference input image is displayed above two interactive 3D viewers labeled A and B. Participants can rotate, zoom, and pan the models, with optional synchronized camera controls, and select either \textit{``A is better''} or \textit{``B is better''} based on their similarity to the reference image and overall visual quality.
}
\label{fig:user_study_interface}
\end{figure}

\subsubsection{Results}

\begin{figure}[t]
\centering
\includegraphics[width=1.0\textwidth]{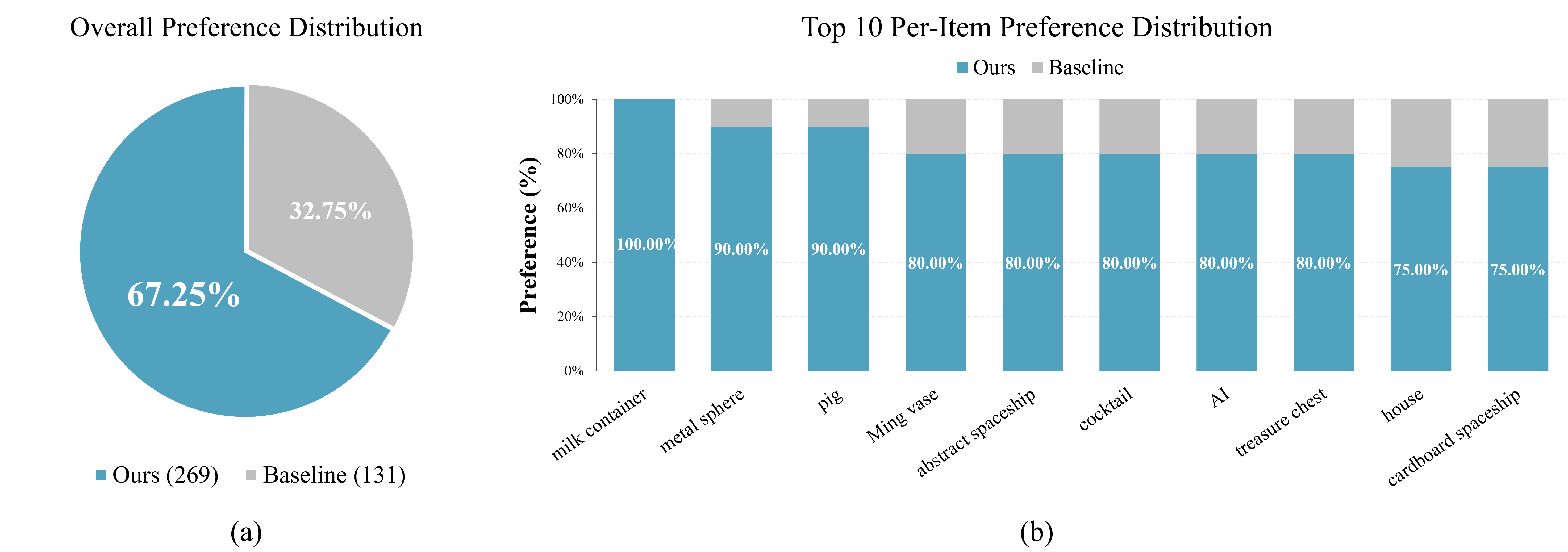}
\caption{
\textbf{Quantitative results of the user study.} (a) Overall preference distribution, where our method is preferred in 67.25\% of the total 400 votes, significantly outperforming the baseline. (b) Top 10 per-item preference histogram, showing the per-sample preference rate for individual test cases. Our approach consistently receives higher user preference across diverse prompts, demonstrating the robustness and superior quality of our 3D asset generation.}
\label{fig:user_result_pie}
\end{figure}

\textbf{Overall Preference.} Aggregated across all 400 pairwise comparisons, participants showed a clear preference for \ourmethod-enhanced outputs, which were selected in \textbf{67.25\%} of trials (269 votes), while the TRELLIS.2 baseline was preferred in only \textbf{32.75\%} of trials (131 votes). As illustrated in Figure~\ref{fig:user_result_pie}, this consistent performance advantage is further illustrated by the per-item distribution, where the adoption from our method significantly outperforms the baseline. 

\textbf{Per-Sample Analysis.} Figure~\ref{fig:user_result_pie} reports the per-sample preference distribution. \ourmethod achieves a majority preference on the majority of test samples, with particularly strong advantages on texture-rich objects (e.g., fabric with woven patterns, engraved metallic surfaces, printed text on objects). 
These samples directly exemplify the detail attenuation problem identified in~\Cref{sec:intro}. 
On a small number of texture-sparse samples where both methods produce visually comparable outputs, preferences are approximately balanced.
This pattern is consistent with the expectation that \ourmethod primarily benefits detail-rich scenarios without degrading quality on texture-uniform inputs.

\section{Additional Ablation Studies}
\label{appendix:appendix_ablation}

\subsection{Ablation of the Tiling Parameter $N$}
\label{app:tile_n_parameter}

We ablate the tiling parameter $N\in\{2,3,4,5,6\}$ to examine its effect on image-to-3D generation quality. Experiments are conducted on the full 3D-Arena~\cite{3d-arena} dataset, comprising 101 images, using TRELLIS.2 with a resolution setting of $1024$. Both global images and local tiles follow the standard preprocessing of the corresponding DINO encoder and are resized to its native input resolution before feature extraction. We use $N=3$ in all main experiments.

Table~\ref{tab:tile_n_parameter} reports the mean scores. Among the evaluated settings, $N=3$ achieves the highest mean PSNR and SSIM and the lowest mean LPIPS. The mean scores are similar for $N=2$--$4$, while $N=5$ and $N=6$ yield lower PSNR and SSIM and higher LPIPS. These results indicate that increasing $N$ does not necessarily improve generation quality and support our default choice of $N=3$.

\begin{table}[t]
\centering
\small
\setlength{\tabcolsep}{9pt}
\renewcommand{\arraystretch}{1.12}
\caption{\textbf{Effect of the tiling parameter $N$ on 3D-Arena.} Bold values indicate the best results among the evaluated settings.}
\label{tab:tile_n_parameter}
\begin{tabular}{lccccc}
\toprule[1pt]
$N$ & PSNR $\uparrow$ & SSIM $\uparrow$ & ULIP-2 $\uparrow$ & Uni3D $\uparrow$ & LPIPS $\downarrow$ \\
\midrule
2 & 21.1799 & 0.8694 & 0.3787 & 0.3469 & 0.4615 \\
\textbf{3 (default)} & \textbf{21.3149} & \textbf{0.8721} & \textbf{0.3837} & \textbf{0.3551} & \textbf{0.4597} \\
4 & 20.9194 & 0.8623 & 0.3805 & 0.3490 & 0.4602 \\
5 & 20.7935 & 0.8557 & 0.3732 & 0.3378 & 0.4643 \\
6 & 20.5157 & 0.8491 & 0.3701 & 0.3349 & 0.4710 \\
\bottomrule[1pt]
\end{tabular}
\end{table}

As shown in Table~\ref{tab:tile_n_parameter}, $N=3$ achieves the highest mean PSNR and SSIM and the lowest mean LPIPS among the evaluated settings. The mean scores remain similar for $N=2$--$4$, whereas larger values yield lower PSNR and SSIM and higher LPIPS. These results support the default choice of $N=3$.

\subsection{Ablation of the Tile Expansion Direction}
\label{app:tile_expansion_direction}

We also examine the effect of tile expansion direction. We keep the same seed, resolution, and sampling steps as above. The \textit{global-to-local} schedule starts with broad tile coverage and progressively concentrates on a smaller set of local tiles, whereas our \textit{local-to-global} schedule initially emphasizes high-priority tiles and gradually broadens the effective tile coverage. Here, the schedule directions refer to local tile coverage; the global condition remains active in both dynamic variants.

\begin{table}[t]
\centering
\small
\setlength{\tabcolsep}{10pt}
\renewcommand{\arraystretch}{1.12}
\caption{\textbf{Effect of tile expansion direction on 3D-Arena.} Higher scores are better for both metrics.}
\label{tab:tile_expansion_direction}
\begin{tabular}{lcc}
\toprule[1pt]
Schedule & ULIP-2 $\uparrow$ & Uni3D $\uparrow$ \\
\midrule
Baseline & 0.3676 & 0.3327 \\
global-to-local & 0.3723 & 0.3406 \\
\rowcolor{gray!12}
local-to-global (ours) & \textbf{0.3837} & \textbf{0.3551} \\
\bottomrule[1pt]
\end{tabular}
\end{table}

As shown in Table~\ref{tab:tile_expansion_direction}, the local-to-global schedule achieves the highest ULIP-2 and Uni3D scores among the evaluated settings. 
These results support the local-to-global expansion used in \ourmethod, which progressively incorporates broader regional evidence while initially emphasizing high-priority local information.

\subsection{Ablation of the Dynamic Conditioning Parameter $\beta$.}
We further examine the effect of the dynamic conditioning parameter $\beta$ on local detail preservation and semantic alignment. In \dycond, $\beta$ controls the onset and slope of the progressive activation schedule, together with the sharpness of rank-based tile weighting.
\Cref{tab:ablation_beta_appendix} reports quantitative results using PSNR to assess image-space fidelity and ULIP-2 and Uni3D to evaluate 3D-aware semantic alignment. Among the evaluated settings, $\beta=2$ achieves the best scores.

\begin{figure*}[t]
\centering
\includegraphics[width=0.999\textwidth]{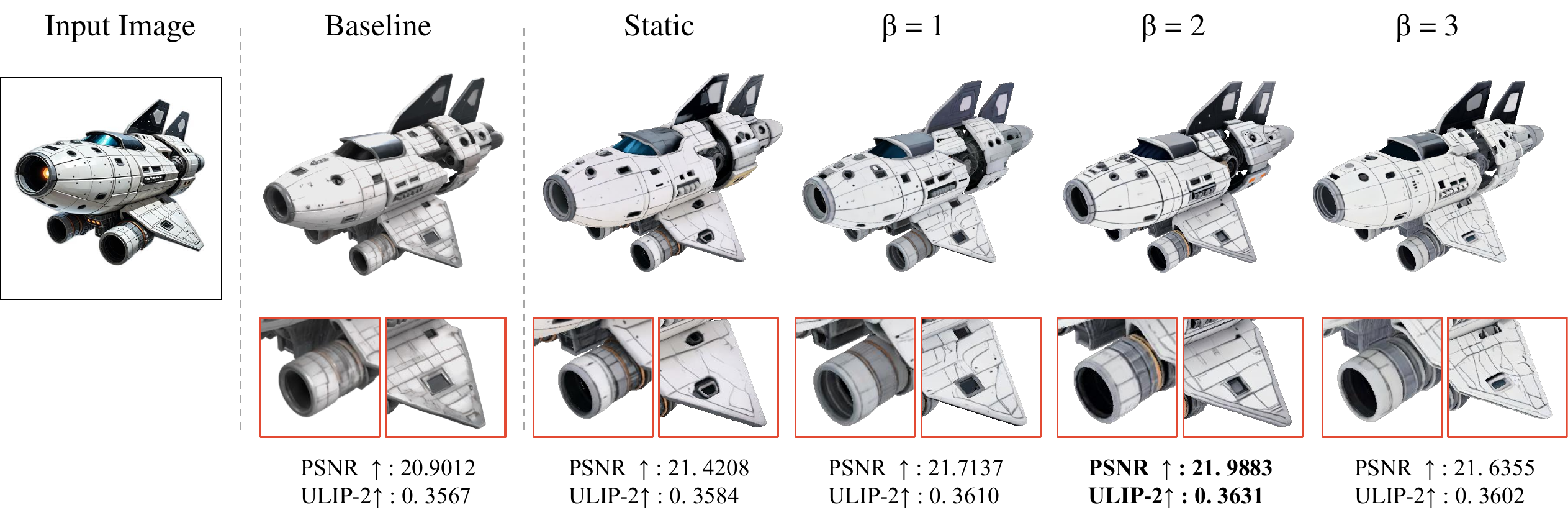}
\caption{
\textbf{Ablation of the dynamic conditioning parameter $\beta$.} From left to right: input image, baseline, static tile conditioning, and results with $\beta=1,2,3$. Red boxes highlight local details. PSNR and ULIP-2 are reported for the illustrated example.}
\label{fig:appendix_ablation_beta}
\end{figure*}

\Cref{fig:appendix_ablation_beta} provides qualitative comparisons between the baseline, static conditioning, and different dynamic settings. The highlighted engine and wing regions illustrate differences in local detail preservation. In this example, all three dynamic settings outperform static conditioning in both PSNR and ULIP-2, with $\beta=2$ achieving the highest scores. These observations support progressive tile activation over fixed tile conditioning and our choice of $\beta=2$.

\begin{table}[t]
\centering
\small
\setlength{\tabcolsep}{9pt}
\renewcommand{\arraystretch}{1.12}
\caption{\textbf{Effect of the dynamic conditioning parameter $\beta$ on 3D-Arena.} Bold values indicate the best reported results.}
\label{tab:ablation_beta_appendix}
\begin{tabular}{lccc}
\toprule[1pt]
Setting & PSNR $\uparrow$ & ULIP-2 $\uparrow$ & Uni3D $\uparrow$ \\
\midrule
Baseline & 20.6741 & 0.3676 & 0.3327 \\
\midrule
Static & 20.9117 & 0.3733 & 0.3446 \\
$\beta$=1 & 21.1564 & 0.3795 & 0.3488 \\
\textbf{$\beta$=2} & \textbf{21.3149} & \textbf{0.3837} & \textbf{0.3551} \\
$\beta$=3 & 21.2213 & 0.3801 & 0.3529 \\
\bottomrule[1pt]
\end{tabular}
\end{table}

\subsection{Ablation of the Fusion Ratio $\alpha$.}
We further examine the effect of the fusion ratio $\alpha$ on local detail preservation and semantic alignment, with $\beta$ fixed at $2$. As the blending coefficient between $f^g$ and $f^{a,t}$, $\alpha$ controls the relative contributions of global guidance and local tile information. 
\Cref{tab:ablation_alpha_appendix} reports results on 3D-Arena, using PSNR, SSIM, and LPIPS to assess image-space fidelity and ULIP-2 and Uni3D to evaluate 3D-aware semantic alignment. Among the evaluated settings, $\alpha=0.4$ achieves the best results.

\begin{table}[t]
\centering
\small
\setlength{\tabcolsep}{9pt}
\renewcommand{\arraystretch}{1.12}
\caption{\textbf{Effect of the fusion ratio $\alpha$ on 3D-Arena.} Bold values indicate the best reported results.}
\label{tab:ablation_alpha_appendix}
\begin{tabular}{lccccc}
\toprule[1pt]
Setting & PSNR $\uparrow$ & SSIM $\uparrow$ & ULIP-2 $\uparrow$ & Uni3D $\uparrow$ & LPIPS $\downarrow$ \\
\midrule
Baseline & 20.6741 & 0.8480 & 0.3676 & 0.3327 & 0.4693 \\
$\alpha$=0.3 & 21.1650 & 0.8696 & 0.3801 & 0.3515 & 0.4655 \\
\textbf{$\alpha$=0.4} & \textbf{21.3149} & \textbf{0.8721} & \textbf{0.3837} & \textbf{0.3551} & \textbf{0.4597} \\
$\alpha$=0.5 & 21.1094 & 0.8693 & 0.3799 & 0.3492 & 0.4649 \\
$\alpha$=0.6 & 20.9245 & 0.8601 & 0.3743 & 0.3416 & 0.4696 \\
\bottomrule[1pt]
\end{tabular}
\end{table}

\begin{figure*}[t]
\centering
\includegraphics[width=0.999\textwidth]{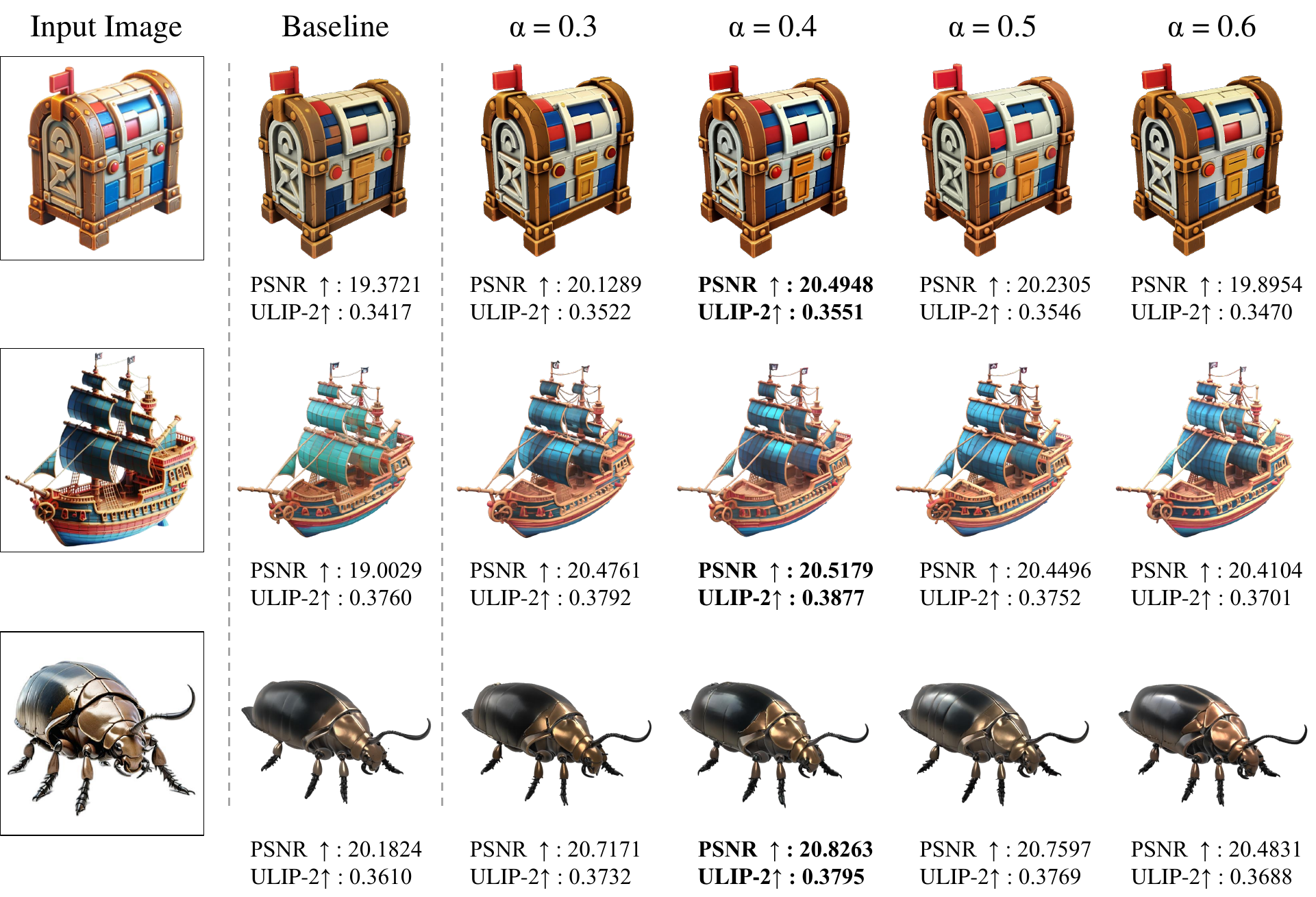}
\caption{\textbf{Ablation of the fusion ratio $\alpha$.} Different fusion ratios affect local detail preservation and consistency with the input image.}
\label{fig:appendix_ablation_alpha}
\end{figure*}

\Cref{fig:appendix_ablation_alpha} provides qualitative comparisons under different settings. An appropriate fusion ratio improves detail preservation while maintaining semantic alignment, whereas overly small or large values may reduce PSNR and, in some cases, also lower ULIP-2. These observations highlight the importance of balancing local conditioning with global guidance.

\section{Discussion and Future Works}
\label{sec:discussion}

\subsection{Discussion}

Evaluating image-to-3D generation remains challenging because its evaluation protocol fundamentally differs from that of conventional 3D reconstruction.
3D reconstruction typically assumes posed input images and follows strict multi-view geometric constraints, where evaluation mainly focuses on how well the reconstructed scene matches the observed views or ground-truth geometry under known camera poses.
By contrast, image-to-3D generation does not require input camera poses and aims to produce a complete 3D asset from a single image, including plausible geometry, texture, and even material for both visible and unseen regions.
Since the generated 3D asset is not guaranteed to be view-aligned with the input image or any predefined ground-truth coordinate frame, the same input image may correspond to multiple plausible 3D assets with different geometry, object scale, canonical orientation, and unseen-region appearance.
Therefore, conventional reconstruction metrics, such as PSNR, SSIM~\cite{wang2004ssim}, Chamfer Distance, F-score, and LPIPS~\cite{zhang2018lpips} on matched-view renderings, measure agreement with a reference under a specific evaluation protocol. Their interpretation should account for alignment conditions and the possibility of multiple plausible 3D outputs, and they should be considered alongside semantic alignment metrics for a comprehensive assessment of image-to-3D generation quality.

Following current image-to-3D generation evaluation metrics, we adopt commonly used metrics from recent works such as TRELLIS~\cite{xiang2025trellis2,xiang2025structured_trellis}, Hunyuan3D~\cite{hunyuan3d22025tencent_v2,hunyuan3d2025hunyuan3d_v21}, and related image-to-3D methods~\cite{feng2025seed3d,li2025step1x_3d,ye2025hi3dgen}. 
Existing evaluation metrics can be broadly grouped into two categories:

The first category evaluates generated assets through rendered 2D views.
In this setting, a 3D asset is rendered into multiple images, and the rendered views are compared with the input image using image-space or feature-space metrics such as CLIP-I~\cite{radford2021clip}, DINO~\cite{caron2021dino}, LPIPS~\cite{zhang2018lpips}, or FID-style distances, as reported in Table~\ref{tab:appendix_all_additional_metrics}.
These metrics directly reflect the visible appearance of the asset, including texture, color, and rendered visual details.
However, they are sensitive to rendering viewpoints, lighting, image resolution, background, and the mismatch between single-view conditioning and multi-view evaluation.

The second category evaluates generated assets from a more 3D-aware representation perspective.
Following recent image-to-3D evaluation protocols, a common practice is to convert the mesh into a colored point cloud and compute image-to-point-cloud similarity using multimodal 3D foundation models such as ULIP-2~\cite{xue2024ulip} and Uni3D~\cite{zhou2023uni3d}.
These metrics provide useful image-to-asset alignment signals and are better suited for object-level semantic and structural consistency than purely rendered-view metrics.
However, they remain primarily alignment-oriented.
Recent 3D generation benchmarks show that 3D asset quality involves multiple dimensions, including geometry details, texture quality, geometry-texture coherence, and prompt-asset alignment~\cite{zhang2025_3dgenbench}.
Therefore, global image-to-point-cloud similarity may overlook fine-grained aspects of texture-focused 3D generation, such as texture fidelity, material boundaries, and surface details.

Overall, these two types of metrics provide complementary but incomplete views of image-to-3D generation quality.
In our evaluation, we therefore combine rendered-view metrics with 3D-aware image-to-point-cloud similarity metrics, and use them together to assess different aspects of 3D generation quality, as shown in Table~\ref{tab:appendix_all_additional_metrics}. 
3DGen-Bench~\cite{zhang2025_3dgenbench} evaluates 3D assets from multiple aspects, including geometry detail, texture quality, geometry coherence, and text--asset consistency. This shows that a single global alignment score is not enough to fully measure 3D generation quality.

Automatic metrics and human preference studies should therefore be viewed as complementary rather than interchangeable: the former enables scalable and reproducible comparisons across methods, while the latter captures perceptual quality, local texture realism, geometry-texture consistency, and human-sensitive detail fidelity.
Accordingly, our evaluation combines automatic metrics with a user study to provide both quantitative and perceptual evidence for the effectiveness of \ourmethod.

\subsection{Future Work}

This work opens up several promising directions for future research.

First, it would be valuable to further explore the broader role of DINO-based visual representations in image-to-3D generation.
Beyond serving as conditioning signals, DINO features may provide useful cues for local correspondence, semantic part awareness, texture placement, and geometry--texture alignment, since they encode both high-level object semantics and spatially localized appearance information.
A deeper understanding of how these features support different stages of 3D generation could inspire more effective local guidance mechanisms and improve the controllability of fine-grained asset generation.

Second, future work may explore more comprehensive evaluation metrics for image-to-3D generation.
Current automated metrics provide useful and reproducible signals for measuring image-to-asset alignment, while human studies offer complementary insight into perceptual quality, texture realism, and geometry--texture consistency.
Developing more detail-sensitive benchmarks, standardized rendering settings, and native 3D evaluation metrics that better reflect human visual perception would further support the evaluation of high-fidelity 3D assets.

Finally, DINO-based representations may also be useful beyond the specific image-to-3D setting studied in this work.
The combination of global and tiled visual features could potentially benefit related tasks that require both holistic semantic understanding and fine-grained local fidelity, such as texture generation, 3D asset editing, image-conditioned refinement, and multi-view content generation.
Exploring these applications may further reveal the general utility of DINO features for controllable and detail-preserving 3D content creation.

\section{Algorithm}
\label{sec:algorithm}

We present \ourmethod\ in a modular form.
Algorithm~\ref{alg:btcemb} constructs the foreground-weighted tile embedding. 
Algorithm~\ref{alg:dycond} computes the timestep-dependent conditioning feature.
Algorithm~\ref{alg:btc3d} integrates both components into the sampling process of a pretrained flow-based generator.

\minisection{Algorithm for \btcembed.}
Given an input image $I$ and its foreground mask $M$, \btcembed\ extracts global and local image features. Each tile is assigned a priority score according to its foreground coverage, and these scores are directly normalized to obtain the static blending weights. The tile features and their weights are jointly sorted for subsequent dynamic conditioning.

Here, $M$ is obtained during input preprocessing and is expressed in the same coordinate system as $I$. \textsc{ImageTiling} returns the image tiles together with their corresponding pixel regions $\{\Omega_k\}_{k=1}^{K}$. The encoder $E$ includes its standard preprocessing, so the full-image and tile features have compatible shapes.
We use $N \geq 2$, with $K=N^2$, and assume $\sum_{k=1}^{K} e_k > 0$.

\minisection{Algorithm for \dycond.}
Given the sorted tile features and static weights, \dycond\ computes timestep-dependent tile weights. Higher-priority tiles receive greater relative emphasis early in sampling, while the relative contributions of lower-priority tiles increase as sampling progresses. The resulting dynamic tile embedding is blended with the global feature using a fixed fusion ratio $\alpha$.

\minisection{Inference pipeline.}
The global and tile features are computed once and reused throughout sampling.
At each sampling step, \dycond\ produces the conditioning feature supplied to the pretrained generator.
Let $t \in [0, 1]$ denote the flow matching trajectory, where $t = 0$ and $t = 1$ represent the source distribution and the target.

\textsc{FlowStep} denotes an update from $t_n$ to $t_{n+1}$ using the backbone's sampling rule.
The pseudocode describes the sampling stage to which \ourmethod\ is applied and returns its generated latent.
All pretrained model parameters remain unchanged.

\clearpage
\newpage

\begin{algorithm}[t]
\caption{Blended Tile Conditioning Embedding (\btcembed)}
\label{alg:btcemb}
\small
\begin{algorithmic}[1]
\Require Input image $I$, foreground mask $M$,
         image encoder $E$, tiling parameter $N \geq 2$
\Ensure Global feature $f^g$,
        sorted tile features $\{f_k^l\}_{k=1}^{K}$,
        sorted static weights $\{w_k\}_{k=1}^{K}$,
        blended tile feature $f^a$

\State $K \gets N^2$
\State $f^g \gets E(I)$
\State $\{(I_k,\Omega_k)\}_{k=1}^{K}
       \gets \Call{ImageTiling}{I,N}$

\For{$k=1$ to $K$}
    \State $f_k^l \gets E(I_k)$
    \State $e_k \gets
        \dfrac{1}{|\Omega_k|}
        \sum_{\mathbf{x}\in\Omega_k} M(\mathbf{x})$
\EndFor

\For{$k=1$ to $K$}
    \State $w_k \gets
        \dfrac{e_k}{\sum_{j=1}^{K}e_j}$
\EndFor

\State $f^a \gets \sum_{k=1}^{K} w_k f_k^l$
\State Jointly sort $\{(f_k^l,w_k)\}_{k=1}^{K}$
       such that $w_1 \geq w_2 \geq \cdots \geq w_K$
\State \Return
       $f^g,\{f_k^l\}_{k=1}^{K},\{w_k\}_{k=1}^{K},f^a$
\end{algorithmic}
\end{algorithm}

\begin{algorithm}[t]
\caption{Dynamic Conditioning (\dycond)}
\label{alg:dycond}
\small
\begin{algorithmic}[1]
\Require Global feature $f^g$,
         sorted tile features $\{f_k^l\}_{k=1}^{K}$,
         sorted static weights $\{w_k\}_{k=1}^{K}$,
         fusion ratio $\alpha \in [0,1]$,
         schedule parameter $\beta \geq 1$,
         timestep $t \in [0,1]$
\Ensure Blended conditioning feature $f^{c,t}$

\State $\lambda(t) \gets
       \mathrm{clip}(\beta t-\beta+1,\,0,\,1)$

\For{$k=1$ to $K$}
    \State $g_k(t) \gets
        \operatorname{sigmoid}\!\left(
        \beta\left(\lambda(t)-\dfrac{k-1}{K-1}\right)
        \right)$
\EndFor

\For{$k=1$ to $K$}
    \State $w_k^t \gets
        \dfrac{g_k(t)\,w_k}
        {\sum_{j=1}^{K}g_j(t)\,w_j}$
\EndFor

\State $f^{a,t} \gets \sum_{k=1}^{K} w_k^t f_k^l$
\State $f^{c,t} \gets
       (1-\alpha)\,f^g+\alpha\,f^{a,t}$
\State \Return $f^{c,t}$
\end{algorithmic}
\end{algorithm}

\begin{algorithm}[t]
\caption{BTC3D-Conditioned Inference}
\label{alg:btc3d}
\small
\begin{algorithmic}[1]
\Require Input image $I$, foreground mask $M$,
         image encoder $E$,
         pretrained flow-based generator $\mathcal{F}_{\theta}$,
         time grid $\{t_n\}_{n=0}^{T}$ with
         $0=t_0<t_1<\cdots<t_T=1$,
         tiling parameter $N \geq 2$,
         fusion ratio $\alpha \in [0,1]$,
         schedule parameter $\beta \geq 1$
\Ensure Generated latent $x_{t_T}$

\State $(f^g,\{f_k^l\}_{k=1}^{K},
        \{w_k\}_{k=1}^{K},f^a)
       \gets \Call{BTCemb}{I,M,E,N}$
\State Sample $x_{t_0} \sim p_{\mathrm{source}}$

\For{$n=0$ to $T-1$}
    \State $f^{c,t_n} \gets
        \Call{DyCond}{f^g,\{f_k^l\}_{k=1}^{K},
        \{w_k\}_{k=1}^{K},\alpha,\beta,t_n}$
    \State $x_{t_{n+1}} \gets
        \Call{FlowStep}{\mathcal{F}_{\theta},
        x_{t_n},f^{c,t_n},t_n,t_{n+1}}$
\EndFor

\State \Return $x_{t_T}$
\end{algorithmic}
\end{algorithm}

\end{document}